\documentclass[letterpaper, 10pt, conference]{ieeeconf}
\IEEEoverridecommandlockouts  
\usepackage[T1]{fontenc}
\usepackage{newtxtext,newtxmath}
\usepackage{amsmath}
\usepackage{graphicx}
\usepackage{booktabs}
\usepackage{multirow}
\usepackage{cite}
\usepackage{tikz}
\usetikzlibrary{positioning, calc}
\usepackage[font=footnotesize]{caption}
\usepackage[font=footnotesize]{subcaption}
\usepackage{hyperref}

\title{\LARGE \bf Image-Space Refraction Correction for Underwater 3D Reconstruction: Warping Flat-Port Views into Pinhole Perspective}

\author{Chelim Lim, Tobias Fischer, Emilio Olivastri, Beverley Gorry, Michael Milford and Alejandro Fontan}

\begin{document}
\thispagestyle{empty}
\pagestyle{empty}

\maketitle
\begin{abstract}
Consumer-grade cameras in flat-port housings are widely used for underwater exploration and mapping of coral reefs and seafloor habitats due to their low cost and accessibility. However, refraction at the flat-port interfaces causes bowl-shaped deformation of the reconstructed scene and camera trajectory, compromising the metric accuracy that mapping and navigation depend on.
To remove the dominant refractive distortion before reconstruction, we introduce a physics-based refraction correction in image-space.
Our method is downstream-agnostic, and the refraction-corrected images can be directly used as input to any reconstruction or SLAM algorithms.
We characterize the refractive distortion through ray-tracing simulations and validate the correction on two real underwater datasets with differing scene structure. 
Compared with conventional and refractive Structure-from-Motion (SfM), our approach removes the reconstruction deformation while registering more frames and maintaining low reprojection error.
The correction further generalizes across diverse reconstruction and VSLAM backends, demonstrating its broad applicability to downstream vision pipelines. Code and interactive demo are available at \href{https://github.com/cllim118/Refrax}{github.com/cllim118/Refrax} and \href{https://huggingface.co/spaces/cllim118/refrax-demo}{huggingface.co/spaces/cllim118/refrax-demo}.
\end{abstract}

\section{Introduction}
\label{sec:introduction}

Recovering accurate 3D geometry underwater is affected by \emph{refraction}, as light bends at the camera–housing interface between media of different refractive indices, such as water, glass, and air, invalidating the single-viewpoint (pinhole) assumption~\cite{Treibitz2012, Chadebecq2020}.

Conventional cameras protected by flat-port housings provide a simple and effective solution for underwater imaging, and are widely used instead of dedicated underwater imaging systems, which are expensive and impractical for many applications~\cite{raoult2016gopros, Singh2024imu, Singh2024stereo}. However, the water--housing interface introduces refractive distortions that violate the perspective (pinhole) assumption, resulting in geometric inaccuracies that accumulate, in 3D applications, into bent, bowl-shaped scene structure and camera trajectory~\cite{She2024, sauder2024scalable}. The resulting inconsistent reconstructions are unsuitable for downstream tasks such as robot localization and navigation, as well as ecological monitoring, including coral reef and benthic habitat surveys~\cite{gorry2025, friedman2026squidle+}.

\begin{figure}[t]
    \centering
    \includegraphics[width=1.0\linewidth]{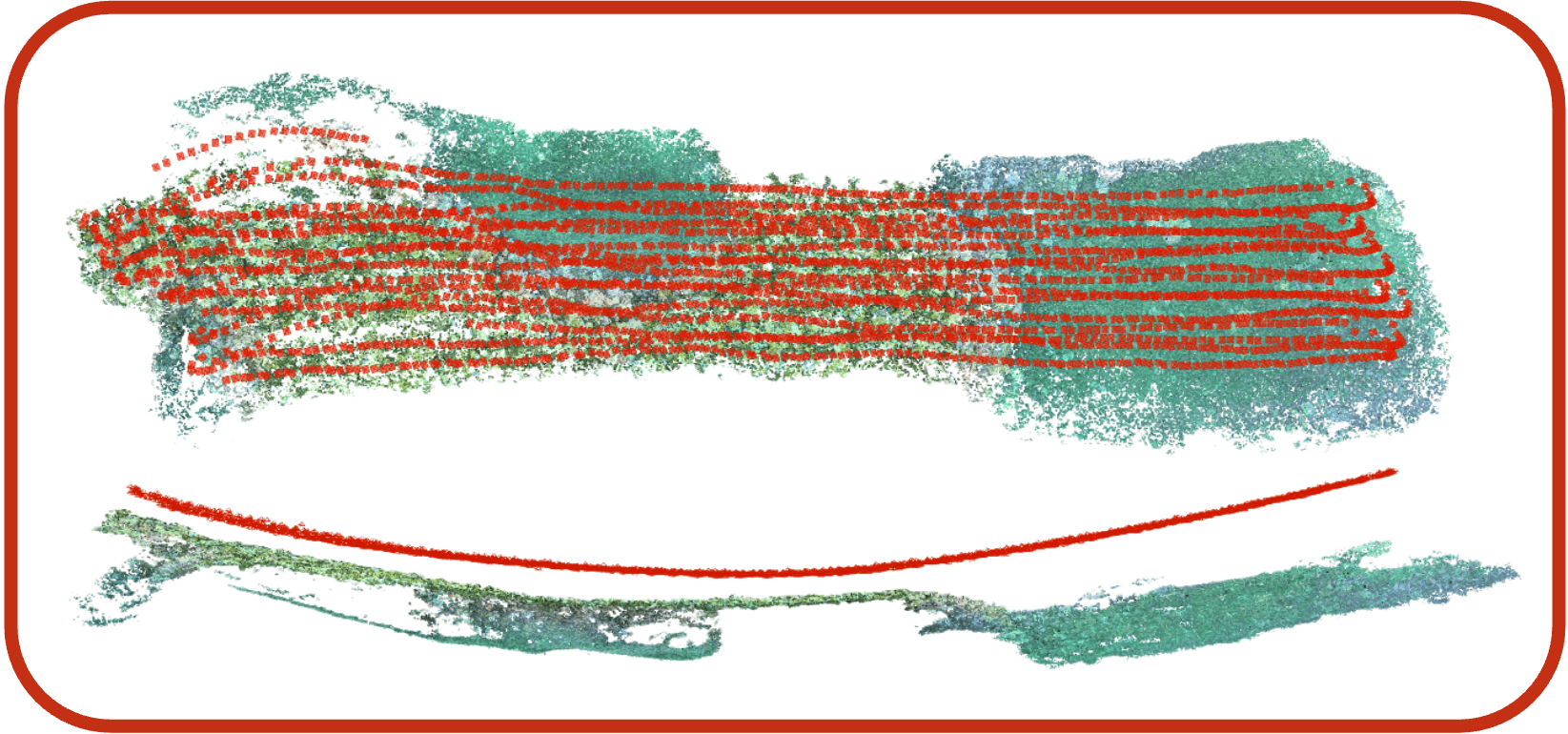}
    \\[0.3em]
    \text{(a) Conventional SfM}

    \vspace{0.8em}

    \includegraphics[width=1.0\linewidth]{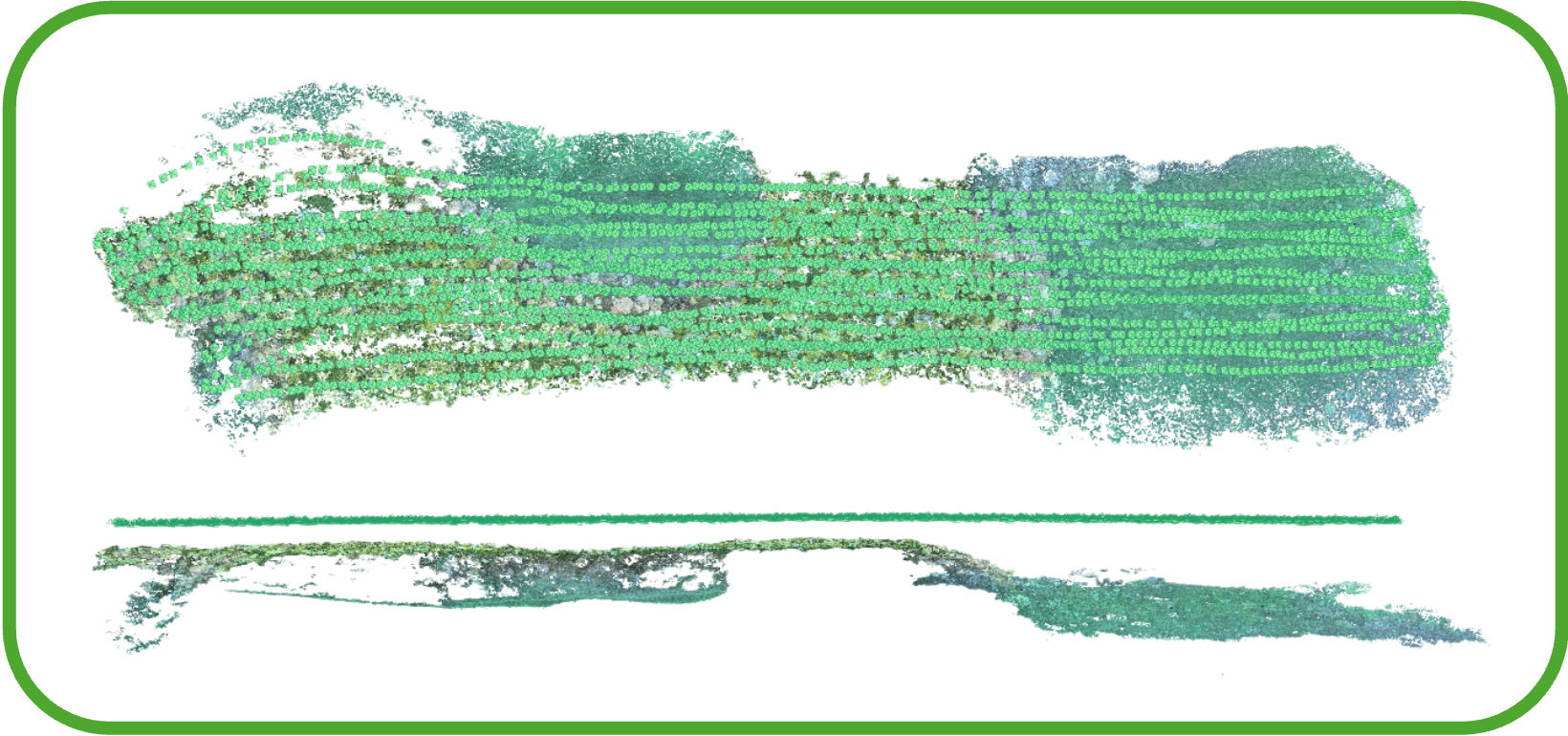}
    \\[0.3em]
    \text{(b) Ours + SfM}

    \caption{
    \textbf{Underwater 3D reconstruction} from top and side views, comparing (a) Conventional SfM with refined intrinsics (red) and (b) Ours~+~SfM (green) on a survey of three coral bommies. Our correction recovers a flat, planar trajectory and scene structure (side view, green), consistent with the vehicle's known survey pattern, whereas Conventional SfM produces a visibly bowed trajectory (side view, red).
    }
    \label{fig:main}
\end{figure}

3D reconstruction methods are mature outside of water~\cite{colmap2016, wang2025vggt, mast3r_eccv24, depthanything3}, but underwater reconstruction still poses challenges, such as how refraction is handled. Two main strategies have addressed this problem. The first absorbs refraction, to a certain extent, into conventional radial-tangential distortion parameters, since the induced distortion has a strong radial component. However, this approximation is not exact, since refraction is not perfectly radial and varies with scene depth. As well as this, estimating these parameters jointly with camera pose and structure during reconstruction hinders optimization convergence and introduces systematic errors.

The second strategy incorporates refraction tightly into the reconstruction through dedicated refractive camera models and modified bundle adjustment~\cite{She2024, Jordt2013}. Such physics-based modeling improves reconstruction accuracy over perspective approximations, but at the cost of requiring refractive parameters to be handled jointly with camera pose and scene structure, which increases the number of coupled variables and hinders optimization convergence. In contrast, we propose to correct refractive distortion in image space, decoupling the correction from the reconstruction, which yields a simpler optimization and makes it independent of the downstream reconstruction system (Fig.~\ref{fig:main}).

The main contribution of this paper is a physics-based refraction correction method that removes the refractive distortion in image space prior to downstream vision tasks. By eliminating the need for refractive parameters from the downstream method, our approach simplifies the optimization and improves geometric fidelity and reconstruction robustness. The corrected images can be used with existing reconstruction, localization, and mapping methods without modifying their camera models or optimization procedures.


In particular, we make the following contributions:
\begin{enumerate}
    \item \textbf{Physics-based preprocessing:}
    A ray-traced image-space correction that characterizes the refractive displacement by tracing light through the flat-port housing's air--glass and glass--water interfaces, and inverts it into a backward warp that maps each image to their equivalent perspective appearance, decoupling refraction correction from the reconstruction backend.
    \item \textbf{Failure-mode analysis:}
    We build a simulator of the flat-port refractive model to analyze the influence of each parameter involved in the warping, and propose several ablations to discuss their relative importance and correlations, and the trade-offs through which over-parameterization hinders optimization convergence and degrades accuracy.
    \item \textbf{Empirical evaluation:}
    On real underwater datasets, our correction improves geometric fidelity by up to $10\times$ over conventional and refractive SfM and transfers across diverse vision pipelines.
\end{enumerate}

\section{Related Work}


Underwater 3D reconstruction faces a wide array of challenges, comprising optical effects such as attenuation and backscattering that degrade image appearance~\cite{watersplatting2025, kweon2026swimm3r}, or sensor fusion with acoustic and inertial sensors for robust localization in turbid scenes~\cite{bagoren2026surfslam, Xu2025aquaslam}. We target the geometric distortion induced by refraction at the camera housing and its impact on 3D reconstruction.


\subsection{Underwater Camera Modeling and Refraction}
\label{sec:camera_modeling}
Underwater cameras enclosed in flat-port housings violate the conventional perspective imaging assumption because light rays are refracted at the interfaces between water, glass, and air. As a result, the observed image rays deviate from those predicted by a pinhole camera model~\cite{Treibitz2012}. This deviation is governed by Snell's law,
\begin{equation}
\eta_w \sin\theta_w = \eta_g \sin\theta_g = \eta_a \sin\theta_a,
\end{equation}
where $\eta_a$, $\eta_g$, $\eta_w$ are the refractive indices of air, glass, and water, and $\theta_a$, $\theta_g$, $\theta_w$ are the ray angles with respect to the interface normal.

Treibitz et al.~\cite{Treibitz2012} showed that a flat-port underwater camera does not have a single viewpoint and cannot be exactly described by a pinhole model, as the required compensation varies with scene depth~\cite{Sedlazeck2012}. In practice, however, the radial component can be partially approximated with conventional camera intrinsics and distortion parameters~\cite{Senshina2022}.

When these parameters are refined through self-calibration, the added degrees of freedom can trade off against one another~\cite{Triggs2000ba}. This is especially problematic in scenes with weak geometric structure, where flat, feature-limited surfaces provide insufficient constraints for reliable self-calibration, which can lead to systematic ``doming''~\cite{griffiths2019comparison, magri2017}. Such deformation is most pronounced on near-planar scenes and reduced on topographically varied ones~\cite{griffiths2019comparison}, and is further amplified under near-parallel viewing~\cite{James2014doming, Wu2014}.

\subsection{Refractive Structure-from-Motion}
\label{sec:refractive_sfm}
Conventional 3D reconstruction pipelines, including COLMAP~\cite{colmap2016}, and recent learning-based approaches such as VGGT~\cite{wang2025vggt}, MASt3R~\cite{mast3r_eccv24}, and Depth Anything 3~\cite{depthanything3} rely on perspective camera models and therefore do not explicitly account for underwater refraction. To address this limitation, refractive SfM methods tightly incorporate light refraction into the reconstruction process to improve geometric accuracy. Chadebecq et al.~\cite{Chadebecq2020} derived an explicit refractive fundamental matrix, extending the theoretical two-view refractive geometry of Chari et al.~\cite{chari2009multiple}, avoiding the costly direct computation of the refractive reprojection error, which requires solving a 12th-degree polynomial. Jordt et al.~\cite{Jordt2013} proposed a complete refractive SfM framework by introducing a refractive reprojection error and customized pose estimation and bundle adjustment procedures. More recently, She et al.~\cite{She2024} integrated refractive geometry into COLMAP, enabling large-scale incremental refractive SfM while jointly optimizing camera and refractive parameters.

All the above-mentioned approaches tightly built refraction into the reconstruction geometry itself -- whether as a refractive two-view formulation~\cite{Chadebecq2020}, a dedicated SfM framework~\cite{Jordt2013}, or an extension of an existing pipeline~\cite{She2024} -- so that the estimation must be reformulated around the refractive model. In contrast, we correct refraction as an image-level pre-processing step, separating it from reconstruction, which enables state-of-the-art reconstruction methods to be used without modification.

\subsection{Camera Calibration Underwater}
\label{sec:calibration}
A complementary line of work addresses underwater refraction without explicitly incorporating refractive geometry into the reconstruction pipeline. Instead, these approaches represent the refractive effect within conventional perspective camera models by adjusting intrinsic parameters or introducing additional correction terms~\cite{Sedlazeck2012, Senshina2022, Lavest2000}. 

The Pinax model~\cite{Luczynski2017pinax} exploits the geometric constraints of practical camera configurations with flat-port housings and represents the refractive projection through a virtual pinhole approximation with precomputed ray mappings. Other approaches estimate physical housing parameters directly from calibration targets~\cite{Seegraber2025calibmar} that can be used by subsequent refractive reconstruction methods~\cite{She2024}. Singh et al.~\cite{Singh2024imu, Singh2024stereo} jointly estimate the refractive index within a visual-inertial framework by decomposing the refractive effect into virtual pinhole and radial distortion components.

Our work complements these approaches by correcting refraction directly in the image domain, as a physics-based preprocessing step before reconstruction. Rather than proposing a new distortion model, we investigate whether correcting refraction improves reconstruction robustness and geometric fidelity while remaining compatible with standard pipelines.

\section{Methodology}

\subsection{Refractive Camera Geometry}
\label{sec:ref_gem_geo}
We first review the geometry of a flat-port underwater camera~\cite{Treibitz2012, Agrawal2012}, which we use to derive the ray-traced displacement field in Sec.~\ref{sec:ref_disp_model}. We consider a pinhole camera enclosed in a flat-port housing observing an underwater scene. Under the conventional perspective model, a 3D point $P = (X,Y,Z) \in \mathbb{R}^3$ projects onto the image plane as
\begin{equation}
u=f_x\frac{X}{Z} + c_x, \qquad
v=f_y\frac{Y}{Z} + c_y,
\end{equation}
where $(u,v)$ is the resulting image coordinate, $(c_x,c_y)$ the principal point, and $(f_x,f_y)$ the focal length. Normalizing by the intrinsics gives the corresponding normalized image coordinates
\begin{equation}
x=\frac{u-c_x}{f_x}, \qquad
y=\frac{v-c_y}{f_y},
\end{equation}
with normalized image radius
\begin{equation}
r=\sqrt{x^2+y^2}.
\end{equation}

For a flat port, the air–glass and glass–water interfaces are assumed parallel to each other and perpendicular to the optical axis~\cite{Treibitz2012, Agrawal2012}. Light refracts at each interface according to Snell's law. The housing is parameterized by the distance $d_0$ between the camera center and the air–glass interface, and the glass thickness $t_g$. With $Z$ denoting the camera-to-object distance along the optical axis, the refractive projection (derived in Sec.~\ref{sec:ref_disp_model}) depends on both $Z$ and the housing geometry. 

\begin{figure}[t!]
\centering

\begin{subfigure}[b]{\columnwidth}
\centering
\resizebox{0.90\columnwidth}{!}{
\begin{tikzpicture}[scale=1.0, >=latex]

    \def\xc{0.0}\def\dz{1.7}\def\tg{0.55}
    \def\xa{\dz}\def\xg{\dz+\tg}\def\Zz{5.3}

    \fill[blue!5]  (\xg,-2.4) rectangle (\Zz,2.4);
    \fill[gray!14] (\xa,-2.4) rectangle (\xg,2.4);
    \node[black] at (0.75,2.15) {\small Air};
    \node[black] at ({(\xg+\Zz)/2},2.15) {\small Water};

    \draw[thick] (\xa,-2.4) -- (\xa,2.4);
    \draw[thick] (\xg,-2.4) -- (\xg,2.4);

    \coordinate (cam) at (\xc,-0.2);
    \fill (cam) circle (2pt);
    \node[left=2pt] at (cam) {\small camera $c$};
    \draw[dotted,gray] (cam) -- (\Zz,-0.2);

    \coordinate (P1) at (\xa,0.85);
    \coordinate (P2) at (\xg,0.98);
    \coordinate (Pw) at (\Zz,2.0);

    \draw[->,thick,red!70!black]     (cam) -- (P1);
    \draw[->,thick,orange!85!black]  (P1)  -- (P2);
    \draw[->,thick,blue!70!black]    (P2)  -- (Pw);

    \node[red!70!black]    at ($(cam)!0.5!(P1)+(-0.15,0.18)$) {\small $\mathbf{r}_a$};
    \node[orange!85!black] at ($(P1)!0.5!(P2)+(0,0.22)$)      {\small $\mathbf{r}_g$};
    \node[blue!70!black]   at ($(P2)!0.5!(Pw)+(-0.1,0.22)$)   {\small $\mathbf{r}_w$};

    \draw[dashed,gray!65] (cam) -- (Pw);

    \fill (P1) circle (1.4pt); \node[below right=0pt] at (P1){\small $P_1$};
    \fill (P2) circle (1.4pt); \node[below right=0pt] at (P2){\small $P_2$};
    \fill[green!50!black] (Pw) circle (2pt);
    \node[right,green!50!black] at (Pw){\small $P_w$};

    \draw[<->] (\xc,-2.05) -- (\xa,-2.05);
    \node[fill=white,inner sep=1pt] at ({\xa/2},-2.05){\small $d_0$};
    \draw[<->] (\xa,-2.05) -- (\xg,-2.05);
    \node[fill=white,inner sep=1pt] at ({\xa+\tg/2},-2.4){\small $t_g$};
    \draw[<->] (\xc,-2.75) -- (\Zz,-2.75);
    \node[fill=white,inner sep=1pt] at ({\Zz/2},-2.75){\small $Z$};

\end{tikzpicture}
}
\caption{\textbf{Ray tracing through the flat-port housing}. A ray from the camera center travels as $\mathbf{r}_a$, refracts toward the normal at the air--glass interface ($P_1$) into $\mathbf{r}_g$, and away from the normal at the glass--water interface ($P_2$) into $\mathbf{r}_w$, meeting the scene plane at depth $Z$ in $P_w$. The dashed line is the direct camera-to-$P_w$ reference. Angles are exaggerated for clarity.}
\label{fig:refraction_geometry}
\end{subfigure}

\vspace{0.6cm}

\begin{subfigure}[b]{\columnwidth}
\centering
\resizebox{0.90\columnwidth}{!}{
\begin{tikzpicture}[
    >=latex,
    every node/.style={inner sep=0}
]
    \def\figwidth{\columnwidth}
    \def\imgwidth{0.34\figwidth}
    \def\dispwidth{0.22\figwidth}

    \node (c) at ({0.5*\imgwidth}, 0)
        {\includegraphics[width=\imgwidth]{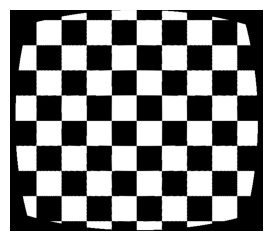}};

    \node (a) at ({\figwidth - 0.5*\imgwidth}, 0)
        {\includegraphics[width=\imgwidth]{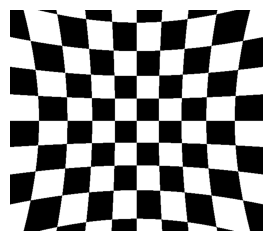}};

    \path (c) -- (a) coordinate[midway] (mid_ca);
    \node (b) at ($(mid_ca) + (0, 0.4\figwidth)$)
        {\includegraphics[width=\imgwidth]{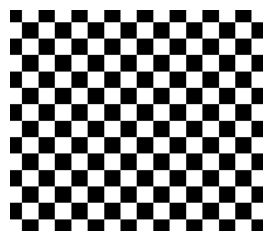}};

    \node (d) at ($(mid_ca) + (0, -0.12\figwidth)$)
        {\includegraphics[width=\dispwidth]{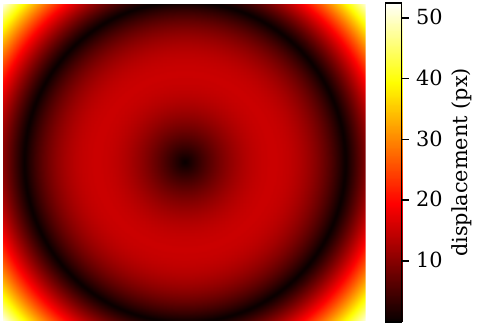}};

    \node[above=1pt of b, font=\small]
        {World 3D Point};

    \node[below=1pt of c, font=\small] (c_label)
        {Virtual Pinhole Image};

    \node[below=1pt of a, font=\small] (a_label)
        {Raw Underwater Image};

    \node[below=1pt of d, font=\scriptsize]
        {Radial Displacement};

    \draw[->,thick]
        (c.north) --
        node[left, xshift=-5pt,font=\scriptsize]
        {back-project}
        (b.south west);

    \draw[->,thick]
        (b.south east) --
        node[right, xshift=5pt,font=\scriptsize]
        {ray tracing}
        (a.north);

    \draw[->,dashed,red!70!black]
        (a_label.south) .. controls
            +(0.0,-0.6) and +(0.0,-0.6) ..
        (c_label.south)
        node[midway, below=10pt, font=\scriptsize, black]
            {bilinear sampling};

\end{tikzpicture}
}
\caption{\textbf{Backward refraction warping}. For each pixel in the corrected image, we solve for the corresponding pixel in the raw underwater image whose ray-traced water-side intersection matches the ideal perspective 3D point at the assumed depth, and sample its color via bilinear interpolation. The radial displacement visualizes the pixel-wise refractive offset between the corrected and raw underwater images. The geometric mapping (solid arrows) proceeds from the corrected image to the 3D point and to the raw image, while color values are fetched in the reverse direction (dashed arrow).}
\label{fig:correction_pipeline}
\end{subfigure}

\vspace{0.6cm}

\begin{subfigure}[b]{\columnwidth}
\includegraphics[width=\linewidth, height=3.4cm, keepaspectratio]{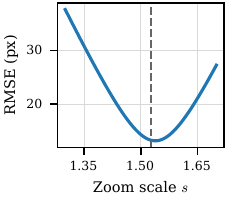}
\includegraphics[width=\linewidth, height=3.4cm, keepaspectratio]{figures/zoom/optimal_zoom.pdf}
\caption{\textbf{Selecting the corrected image's scale factor $s$.} Left: Displacement-field RMSE as a function of the scale factor $s$; the minimum defines $s^*$. Right: Corrected image with the optimal scale $s^*$, closest to the original field of view.}
\label{fig:zoom_optimal}
\end{subfigure}

\caption{\textbf{Refraction-aware underwater image correction.}}
\label{fig:refraction_combined}
\end{figure}

\begin{figure*}[t]
    \centering
    \newcommand{\sensplaceholder}{%
        \fbox{\parbox[c][2.6cm][c]{0.30\linewidth}{\centering\footnotesize Placeholder\\sensitivity plot}}%
    }
    \begin{tabular}{@{}c@{\hspace{0.01\linewidth}}c@{\hspace{0.01\linewidth}}c@{\hspace{0.01\linewidth}}c@{\hspace{0.01\linewidth}}c@{\hspace{0.01\linewidth}}c@{}}
        \includegraphics[width=0.155\linewidth]{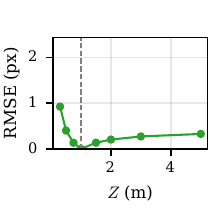} &
        \includegraphics[width=0.155\linewidth]{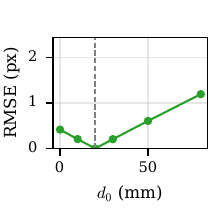} &
        \includegraphics[width=0.155\linewidth]{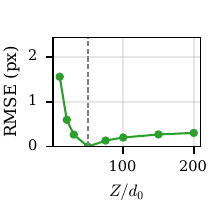} &
        \includegraphics[width=0.155\linewidth]{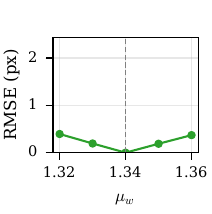} &
        \includegraphics[width=0.155\linewidth]{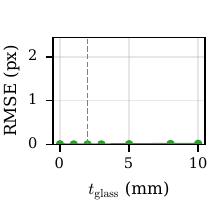} &
        \includegraphics[width=0.155\linewidth]{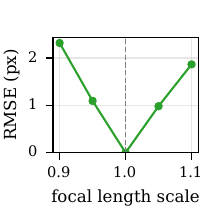} \\[0.1em]
        \footnotesize (a) & \footnotesize (b) & \footnotesize (c) & \footnotesize (d) & \footnotesize (e) & \footnotesize (f) 
    \end{tabular}
    \caption{
        \textbf{Sensitivity analysis of refraction-correction parameters.}
        Each panel varies one parameter of the flat-port housing model while holding the others fixed at their nominal values: (a) depth $Z$, (b) housing-to-camera distance $d_0$, (c) the ratio $Z/d_0$ of depth to housing-to-camera distance, (d) water refractive index $\eta_w$, (e) glass thickness $t_g$, and (f) focal length $f_x,f_y$.
    }
    \label{fig:sensitivity}
\end{figure*}

\subsection{Forward Refractive Displacement Model}
\label{sec:ref_disp_model}
We compute the refractive displacement by tracing rays through the flat-port housing (see Fig.~\ref{fig:refraction_geometry}). Given a pixel $(u,v)$, and its normalized image coordinates $(x,y)$ and radius $r$, as defined in Sec.~\ref{sec:ref_gem_geo}, its ray in air is then back-projected as
\begin{equation}
\mathbf{r}_a = \frac{(x,\,y,\,1)^\top}{\sqrt{r^2+1}} \in \mathbb{S}^2.
\end{equation}
This ray intersects the air--glass interface at $P_1$, located at distance $d_0$ from the camera center along the port normal $\mathbf{n} \in \mathbb{S}^2$. Applying Snell's law gives the ray direction in glass,
\begin{equation}
\mathbf{r}_g =
\operatorname{refract}(\mathbf{r}_a,\mathbf{n},\eta_a,\eta_g),
\end{equation}
where $\eta_a$ and $\eta_g$ are the refractive indices of air and glass. The function $\operatorname{refract}(\cdot)$ applies Snell's law in vector form:
\begin{equation}
\operatorname{refract}(\mathbf{r},\mathbf{n},\eta_1,\eta_2)
= \mu\mathbf{r} + \left(\sqrt{1-\mu^2(1-c^2)} - \mu c\right)\mathbf{n},
\end{equation}
where $\mu=\eta_1/\eta_2$ and $c=\mathbf{n}^\top\mathbf{r}$. The refracted ray propagates through the glass of thickness $t_g$ to the glass--water interface at $P_2$, where a second refraction gives the water-side direction,
\begin{equation}
\mathbf{r}_w =
\operatorname{refract}(\mathbf{r}_g,\mathbf{n},\eta_g,\eta_w),
\end{equation}
with $\eta_w$ denoting the refractive index of water. 

The interface points $P_1$, $P_2$, and the ray directions are determined by the housing geometry and are independent of scene depth, and are given by
\begin{equation}
P_1 = \frac{d_0}{\mathbf{n}^\top \mathbf{r}_a}\,\mathbf{r}_a, \qquad
P_2 = P_1 + \frac{t_g}{\mathbf{n}^\top \mathbf{r}_g}\,\mathbf{r}_g.
\end{equation}
Here and throughout, subscripts $x,y,z$ denote a vector's components along the camera's optical axis and image-plane directions. The water-side ray intersects the scene plane at
\begin{equation}
\label{eq:Pw}
P_w = P_2 + \frac{Z-P_{2,z}}{r_{w,z}}\,\mathbf{r}_w,
\end{equation}
so that $P_{w,z} = Z$ by construction. The corresponding refracted image location is then
\begin{equation}
\pi_r(u,v;Z) =
\left(
f_x\frac{P_{w,x}}{Z}+c_x,\;
f_y\frac{P_{w,y}}{Z}+c_y
\right).
\end{equation}
The refractive displacement is defined as
\begin{equation}
\label{eq:displacement}
\Delta\mathbf{d}(u,v;Z) = \pi_r(u,v;Z)-(u,v).
\end{equation}

\subsection{Backward Refraction Warping}
To remove refraction, we invert the displacement field into a backward resampling map that relocates each refracted pixel toward its perspective position, and use it to warp the input image (see Fig.~\ref{fig:correction_pipeline}). 

\textbf{The housing magnifies the scene}: since water is optically denser than air, rays bend toward the optical axis as they cross the interfaces, so the housing captures a narrower field of view than an equivalent perspective camera of the same focal length~\cite{Treibitz2012}. Consequently, mapping raw pixels back to their perspective positions concentrates the captured content into a smaller radius around the image center, strongly shrinking the corrected image.
To compensate, we introduce a scale factor $s$ that rescales the focal length used in the corrected image's projection, $f_x' = s f_x$ and $f_y' = s f_y$, and resize the output image to new dimensions $(w',h')$ accordingly. We estimate $s$ once per housing configuration via the ray-tracing simulation of Sec.~\ref{sec:ref_disp_model}: we compute the refractive displacement field $\Delta\mathbf{d}(u,v;Z)$ over a range of candidate scales and select the value of $s$ that minimizes its RMSE, keeping the corrected image as close as possible in scale to the original, see Fig.~\ref{fig:zoom_optimal}.

Given this scale factor, we compute the backward correspondence for each output pixel as follows. For each pixel $(u^*,v^*)$ in the corrected image, at known depth $Z^*$, we identify the corresponding pixel $(u,v)$ in the raw underwater image and sample its intensity there via bilinear interpolation. We first back-project $(u^*,v^*,Z^*)$ to its 3D point under the ideal perspective model, using the corrected intrinsics $f_x', f_y'$,
\begin{equation}
P^* = \left(\frac{(u^*-c_x)Z^*}{f_x'},\; \frac{(v^*-c_y)Z^*}{f_y'},\; Z^*\right),
\end{equation}
and then solve for the raw pixel $(u,v)$ whose ray-traced water-side intersection $P_w(u,v;Z^*)$, as defined in Eq.~\eqref{eq:Pw}, coincides with $P^*$, i.e.,
\begin{equation}
(u,v) = \operatorname*{arg\,min}_{(u,v)} \left\lVert P_w(u,v;Z^*) - P^* \right\rVert^2,
\end{equation}
solved as nonlinear least-squares. The intensity at $(u^*,v^*)$ in the corrected image is then set to the bilinearly interpolated intensity of the raw image at $(u,v)$.

\subsection{Depth Assumption in Practice}
While the formulation above supports a per-pixel depth $Z$, our experiments assume a single constant depth $Z_0$ for all images in a sequence, so that the backward mapping is computed once and applied uniformly rather than solved independently for every pixel. We set $Z_0$ to average scene depth over the sequence. This is a reasonable approximation for our datasets, which, as in most coral reef surveys, are largely planar or captured at a roughly constant altitude.

Depth variability nonetheless affects correction quality: as our results on the Lizard Island dataset show, scenes with substantial depth variation are only approximately corrected under a single $Z_0$. Future work should address variable depth, per image or even per pixel, to achieve maximally accurate and reliable reconstructions.

\subsection{Simulation and Sensitivity Analysis}
\label{subsec:sensitivity}
We simulate the refractive displacement (Eq.~\ref{eq:displacement}) while varying each parameter of the flat-port housing model independently, and report the resulting RMSE relative to the nominal configuration (Fig.~\ref{fig:sensitivity}). The displacement is most sensitive to the ratio between the scene depth $Z$ and the housing-to-camera distance $d_0$: varying $Z$ or $d_0$ independently produces substantial RMSE changes, and expressing this dependence through the ratio $Z/d_0$ captures the dominant trend. In contrast, glass thickness $t_g$ has a negligible effect on the displacement, indicating that accurately estimating $Z/d_0$ is far more critical for the correction than precisely measuring the housing's glass thickness.

\section{Experimental Setup}

\label{sec:experimentalsetup}
\subsection{Baselines}
We compare our approach against SfM COLMAP~\cite{colmap2016} and its refractive variant, Refractive COLMAP~\cite{She2024}. COLMAP is the standard perspective baseline that can approximate refraction with a radial-tangential distortion model, while Refractive COLMAP jointly optimizes refractive parameters with camera poses and structure throughout the SfM pipeline. Ours instead applies our physics-based correction to the images before reconstructing with conventional COLMAP. Comparing these three approaches highlights their caveats and benefits in terms of final geometric accuracy, optimization convergence, and computational performance.

Because our correction is applied directly to the image, prior to any reconstruction step, it is agnostic to the specific downstream pipeline. We test this portability by evaluating two learning-based multi-view models, MapAnything~\cite{keetha2026mapanything} and Depth Anything 3 (DA3)~\cite{depthanything3}, on the raw and corrected images. Furthermore, we evaluate five VSLAM systems spanning a range of paradigms~\cite{fontan2025vslam}: the traditional hand-crafted feature-based ORB-SLAM3~\cite{campos2021orb}; the learned feature-based AnyFeature-VSLAM~\cite{fontan2024anyfeature}; the RNN-based DROID-SLAM~\cite{teed2021droid} and DPVO~\cite{lipson2024deep}; and the transformer-based MASt3R-SLAM~\cite{murai2025mast3r}.

\subsection{Datasets}
We evaluate on two datasets of real underwater data captured with consumer GoPro cameras in standard flat-port housings. These were collected by two different teams across four missions at two geographic locations, Malaysia and Lizard Island, spanning four different times of year. Frames from the two datasets are shown in Fig.~\ref{fig:datasets}. Table~\ref{tab:datasets} summarizes the complementary characteristics of both datasets, contrasting diver-handheld versus robot-mounted capture and near-planar versus topographically varied scene geometry. 

\newcommand{\raiseamt}{0.4em}

\begin{figure}[t]
\resizebox{1.0\columnwidth}{!}{
    \centering
    \begin{tabular}{c c c c}
        \raisebox{\raiseamt}{\includegraphics[width=0.48\linewidth]{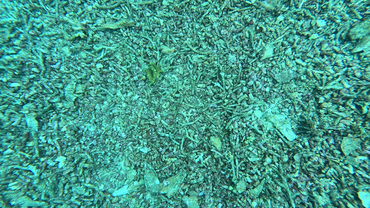}}
        &\includegraphics[width=0.48\linewidth]{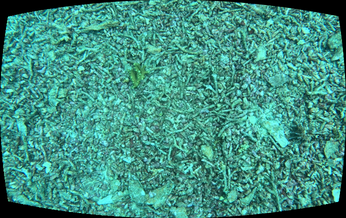}
        &\raisebox{\raiseamt}{\includegraphics[width=0.48\linewidth]{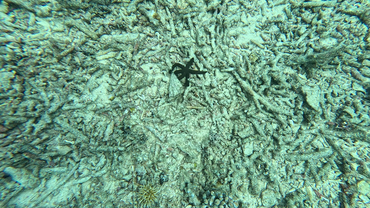}}
        &\includegraphics[width=0.48\linewidth]{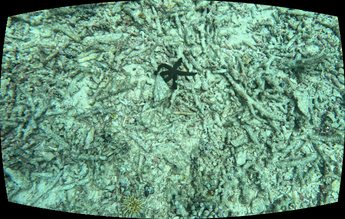}
        \\
        \raisebox{\raiseamt}{\includegraphics[width=0.48\linewidth]{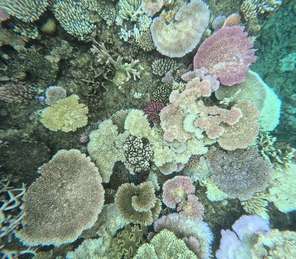}}
        & \includegraphics[width=0.48\linewidth]{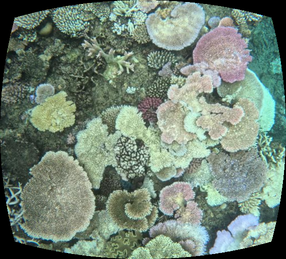}
        & \raisebox{\raiseamt}{\includegraphics[width=0.48\linewidth]{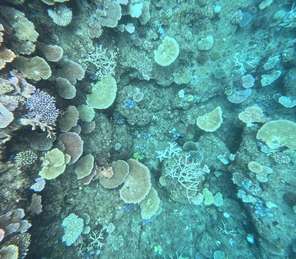}}
        & \includegraphics[width=0.48\linewidth]{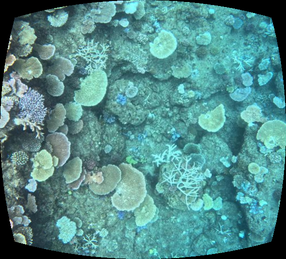}
    \end{tabular}
}
    \caption{\textbf{Top row}: two RGB images from the Malaysia dataset; \textbf{bottom row}: two RGB images from the Lizard Island dataset. In each pair, left is the raw underwater capture and right is the refraction-corrected image.}
    \label{fig:datasets}
\end{figure}





\begin{table}[t]
\centering
\caption{\textbf{Per-sequence characteristics of the Malaysia and Lizard Island datasets.} $Z$ denotes the range of camera-to-scene distance observed in each sequence.}
\label{tab:datasets}
\resizebox{\columnwidth}{!}{%
\begin{tabular}{llcc|c}
\toprule
\textbf{Dataset} & \textbf{Sequence} & \textbf{Frames} & \textbf{$Z$ (m)} & \textbf{Characteristics} \\
\midrule
\multirow{5}{*}{Malaysia}
 & p1~s01 & 17568 & 0.5 & \multirow{5}{*}{\shortstack{Diver handheld \\ GoPro HERO13 \\ Coral rubble \\ $5{\times}5$\,m \\ Near-planar scene}} \\
 & p1~s02 & 7557 & 1.0 & \\
 & p1~s03 & 11043 & 1.5 & \\
 & p3~s02 & 9885 & 1.0 & \\
 & p3~s03 & 5127 & 1.5 & \\
\midrule
\multirow{5}{*}{Lizard Island}
 & Feb~2024~01 & 10473 & \multirow{5}{*}{1--8} & \multirow{5}{*}{\shortstack{ASV \\ GoPro HERO11 \\ Coral reef \\ $57{\times}45$\,m \\ Varied scene}} \\
 & Mar~2024~01 & 11828 & & \\
 & Mar~2024~02 & 4678  & & \\
 & Sep~2024~01 & 9156  & & \\
 & Sep~2024~02 & 14382 & & \\
\bottomrule
\end{tabular}%
}
\end{table}


\textbf{Malaysia Dataset.}~\cite{leung2025decision}
The survey was conducted at Pom Pom Island, Borneo, Malaysia, as part of an initiative evaluating biodegradable biomaterials for coral-rubble stabilization. A diver-handheld, 3-camera GoPro HERO13 stereo rig surveyed a coral-rubble seabed, with camera-to-seafloor distance varying across passes.

\textbf{Lizard Island Dataset.}~\cite{mou2022reconfigurable}
An ASV-mounted GoPro HERO11 over a coral reef. The data was collected as part of a multi-year autonomous reef restoration program on the Great Barrier Reef, using a reconfigurable surface vehicle fleet. The scene is topographically varied, while the ASV maintains a fixed height above the water surface, producing an approximately planar camera trajectory.

\subsection{Evaluation Metrics}
Evaluating underwater 3D reconstruction against absolute ground truth remains impractical at the scale of our surveys. GPS does not propagate underwater, so ASV positioning is not accurate enough to serve as a reliable pose reference, and more sophisticated positioning or scanning systems capable of providing reliable ground truth are expensive and impractical to deploy in the field. We instead evaluate each reconstruction using a thorough combination of structural consistency, whether the reconstruction preserves the underlying geometry known a priori, and standard SfM statistics~\cite{colmap2016, pan2024global, wang2025structure}.

\textbf{Structural deformation.}
A hallmark of refractive distortion is systematic bowl-shaped deformation in reconstructed scene geometry and camera trajectories. We quantify this deformation using a sequence-specific planar reference. For the near-planar Malaysia seafloor, we fit a plane to the reconstructed 3D points, while for Lizard Island we fit a plane to the reconstructed camera centers, reflecting its constant-altitude ASV trajectory.

\emph{Plane fit and scale normalization.}
Given reconstructed points $\{\mathbf{x}_i\}_{i=1}^{N}$, we estimate the best-fit plane by SVD and compute the signed point-to-plane residual
\begin{equation}
r_i = \mathbf{n}^{\top}\mathbf{x}_i + d,
\label{eq:plane_residual}
\end{equation}
where $\mathbf{n}$ is the unit plane normal and $d$ is the plane offset. We orient $\mathbf{n}$ toward the mean camera center. Let $(u_i, v_i)$ denote the coordinates of the in-plane projection of $\mathbf{x}_i$ in an orthonormal basis of the fitted plane. To account for arbitrary monocular scale, we normalize by the projected scene extent
\begin{equation}
D = \sqrt{(\max_i u_i - \min_i u_i)^2 + (\max_i v_i - \min_i v_i)^2}.
\label{eq:scale_norm}
\end{equation}

\emph{Radial curvature.}
We model the systematic bowl/dome component of the residual field with a radially symmetric paraboloid,
\begin{equation}
r_i \approx c_0 + c_1 u_i + c_2 v_i + c_3(u_i^2 + v_i^2),
\label{eq:radial_curvature}
\end{equation}
where the affine terms absorb residual offset and tilt, while $c_3$ captures the systematic radial curvature. We report the dimensionless curvature
\begin{equation}
\kappa = c_3 D,
\label{eq:kappa}
\end{equation}
where $\kappa = 0$ indicates a flat reconstruction and $|\kappa|$ measures the strength of the radial bowl/dome deformation. The sign indicates the deformation direction, with negative values corresponding to a bowl and positive values to a dome.





\newcommand{\tbox}[1]{{\setlength{\fboxsep}{1pt}\boxed{#1}}}

\begin{table}[t!]
\centering
\renewcommand{\arraystretch}{0.95}
\caption{\textbf{Reconstruction results}. Distortion parameters are fixed (\emph{fix}) or refined (\emph{ref}); planarity is measured against the scene (Malaysia) or trajectory (Lizard Island). $\kappa = 0$ is flat; negative is a bowl, positive a dome. Parentheses: insufficient sequence coverage; —$^*$: failed reconstruction.}
\label{tab:main_results}

\centering
\textbf{Lizard Island}\\[2pt]
\resizebox{\linewidth}{!}{%
\begin{tabular}{lllccccc}
\toprule
Sequence & Method & Dist. & Reg.$\uparrow$ & Pts$^{\times 10^5}\uparrow$ & Track$\uparrow$ & Reproj.$\downarrow$ & $\kappa \approx 0$ \\
\midrule
\multirow{3}{*}{feb\_gp1}
& COLMAP & ref & 0.88 & 4.22 & 6.78 & 0.35 & $-0.366$ \\
& Refractive COLMAP & fix & 0.88 & 4.22 & 6.77 & 0.36 & $-0.453$ \\
& Ours + COLMAP & fix & \underline{0.96} & \underline{5.85} & \underline{7.75} & \underline{0.34} &  $\tbox{\mathbf{+0.017}}$ \\
\midrule
\multirow{3}{*}{mar\_gp1}
& COLMAP & ref & 0.99 & 6.72 & 8.35 & 0.48 & $-0.304$\\
& Refractive COLMAP & fix & 0.99 & 6.68 & 8.34 & 0.48 & $+0.055$ \\
& Ours + COLMAP & fix & \underline{1.00} & \underline{8.15} & \underline{9.41} & \underline{0.45} & $\tbox{\mathbf{+0.014}}$ \\
\midrule
\multirow{3}{*}{mar\_gp2}
& COLMAP & ref & 0.90 & 5.99 & 7.90 & 0.48 & $+0.306$ \\
& Refractive COLMAP & fix & 0.90 & 5.98 & 7.88 & 0.48 & $-0.057$ \\
& Ours + COLMAP & fix & \underline{1.00} & \underline{6.99} & \underline{9.25} & \underline{0.47} & $\tbox{\mathbf{+0.017}}$ \\
\midrule
\multirow{3}{*}{sep\_gp1}
& COLMAP & ref & 0.79 & 5.10 & 9.19 & 0.57 & $+0.082$ \\
& Refractive COLMAP & fix & 0.79 & 5.10 & 9.16 & 0.57 & $-0.045$ \\
& Ours + COLMAP & fix & \underline{0.97} & \underline{6.93} & \underline{9.94} & \underline{0.52} & $\tbox{\mathbf{-0.024}}$ \\
\midrule
\multirow{3}{*}{sep\_gp2}
& COLMAP & ref & 0.39 & (3.41) & (8.54) & (0.56) & ($-0.029$) \\
& Refractive COLMAP & fix & 0.39 & (3.41) & (8.52) & (0.56) & ($+0.068$) \\
& Ours + COLMAP & fix & \underline{0.83} & \underline{6.79} & \underline{8.84} & \underline{0.49} & $\tbox{\mathbf{+0.053}}$ \\
\bottomrule
\end{tabular}%
}

\vspace{8pt}

\centering
\textbf{Malaysia}\\[2pt]
\resizebox{\linewidth}{!}{%
\begin{tabular}{lllccccc}
\toprule
Sequence & Method & Dist. & Reg.$\uparrow$ & Pts$^{\times 10^5}\uparrow$ & Track$\uparrow$ & Reproj.$\downarrow$ & $\kappa \approx 0$ \\
\midrule
\multirow{3}{*}{p1\_s01}
& COLMAP & ref & 0.95 & 23.32 & 5.70 & \underline{0.37} & $+0.188$ \\
& Refractive COLMAP & fix & \multicolumn{5}{c}{—$^*$} \\
& Ours + COLMAP & fix & \underline{0.99} & \underline{27.37} &  \underline{6.01} & \underline{0.37} & $\tbox{\mathbf{-0.068}}$ \\
\midrule
\multirow{3}{*}{p1\_s02}
& COLMAP & ref & 1.00 & 12.70 & 7.38 & \underline{0.29} & $+0.089$ \\
& Refractive COLMAP & fix & 1.00 & 12.82 & 7.19 & 0.57 & $+0.694$ \\
& Ours + COLMAP & fix & 1.00 & \underline{13.14} & \underline{8.57} & 0.32 & $\tbox{\mathbf{-0.000}}$ \\
\midrule
\multirow{3}{*}{p1\_s03}
& COLMAP & ref & 1.00 & 11.10 & 9.43 & \underline{0.36} & $+0.073$ \\
& Refractive COLMAP & fix & 1.00 & \underline{11.41} & 8.94 & 0.68 & $+0.105$ \\
& Ours + COLMAP & fix & 1.00 & 10.61 & \underline{12.01} & 0.40 & $\tbox{\mathbf{+0.031}}$ \\
\midrule
\multirow{3}{*}{p3\_s02}
& COLMAP & ref & 1.00 & 12.97 & 5.03 & \underline{0.41} & $-0.064$ \\
& Refractive COLMAP & fix & \multicolumn{5}{c}{—$^*$} \\
& Ours + COLMAP & fix & 1.00 &  \underline{15.56} & \underline{5.40} & 0.42 & $\tbox{\mathbf{-0.004}}$ \\
\midrule
\multirow{3}{*}{p3\_s03}
& COLMAP & ref & 1.00 & 7.30 & 7.45 & \underline{0.35} & $+0.097$ \\
& Refractive COLMAP & fix & 1.00 & 7.32 & 7.34 & 0.62 & $+0.253$ \\
& Ours + COLMAP & fix & 1.00 & \underline{7.76} & \underline{8.73} & 0.38 &$\tbox{\mathbf{+0.078}}$ \\
\bottomrule
\end{tabular}%
}
\end{table}
\begin{figure}[t!]
    \centering

    \newcommand{\lizardimgtop}[1]{%
        \includegraphics[trim={0pt 0pt 0pt 150pt}, clip, width=0.47\linewidth]{#1}%
    }
    \newcommand{\lizardimgbot}[1]{%
        \includegraphics[trim={0pt 180pt 0pt 250pt}, clip, width=0.47\linewidth]{#1}%
    }
    \newcommand{\lizardpair}[2]{%
        \begin{tabular}{@{}c@{}}
            \lizardimgtop{#1}\\[-1.3em]
            \lizardimgbot{#2}
        \end{tabular}%
    }
    \newcommand{\lizardlabel}[1]{%
        \raisebox{-0.25cm}{\rotatebox{90}{\scriptsize #1}}%
    }

    \newcommand{\malaysiaimgtop}[1]{%
        \includegraphics[trim={300pt 50pt 200pt 10pt}, clip, width=0.5\linewidth]{#1}%
    }
    \newcommand{\malaysiaimgbot}[1]{%
        \includegraphics[trim={300pt 250pt 200pt 400pt}, clip, width=0.5\linewidth]{#1}%
    }
    \newcommand{\malaysiapair}[2]{%
        \begin{tabular}{@{}c@{}}
            \malaysiaimgtop{#1}\\[0.3em]
            \malaysiaimgbot{#2}
        \end{tabular}%
    }
    \newcommand{\malaysialabel}[1]{%
        \raisebox{-0.25cm}{\rotatebox{90}{\scriptsize #1}}%
    }

    \begin{subfigure}[t]{\linewidth}
        \centering
        \caption*{\textbf{Lizard-Island}}
        \begin{tabular}{@{}c@{\hspace{0.01\linewidth}}c@{\hspace{0.02\linewidth}}c@{}}
            \lizardlabel{Feb'24} &
            \lizardpair{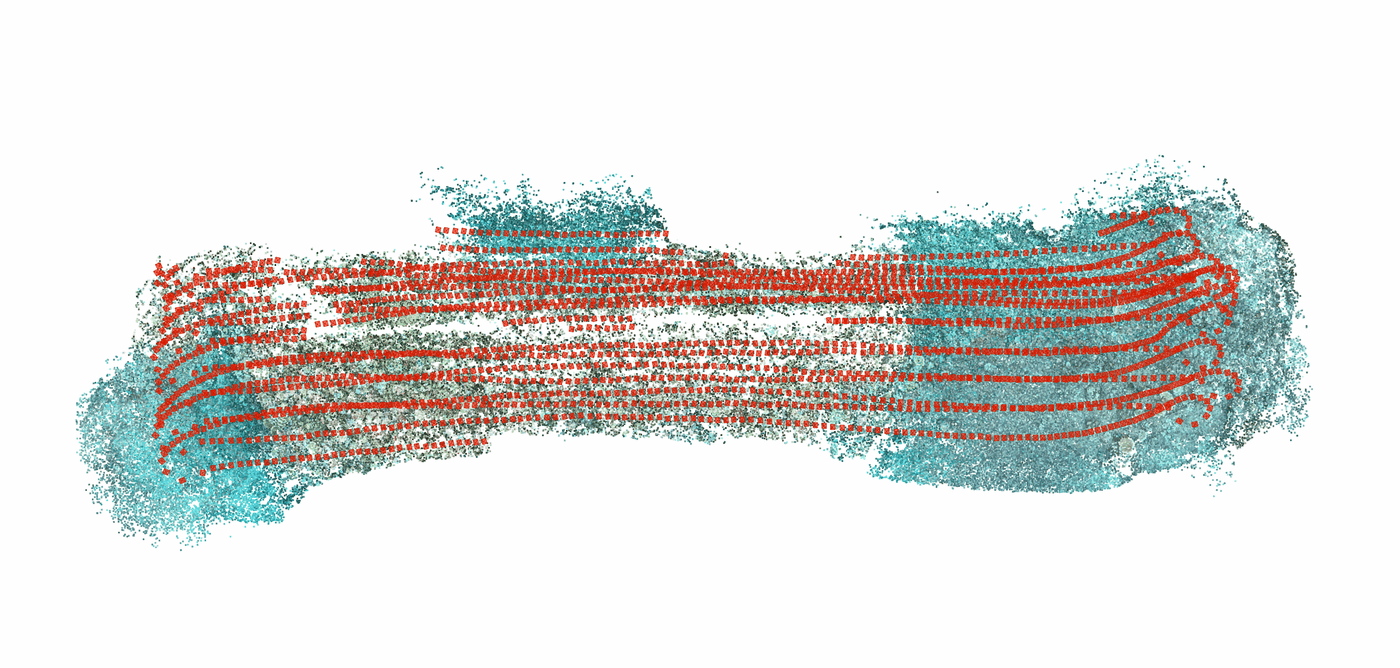}{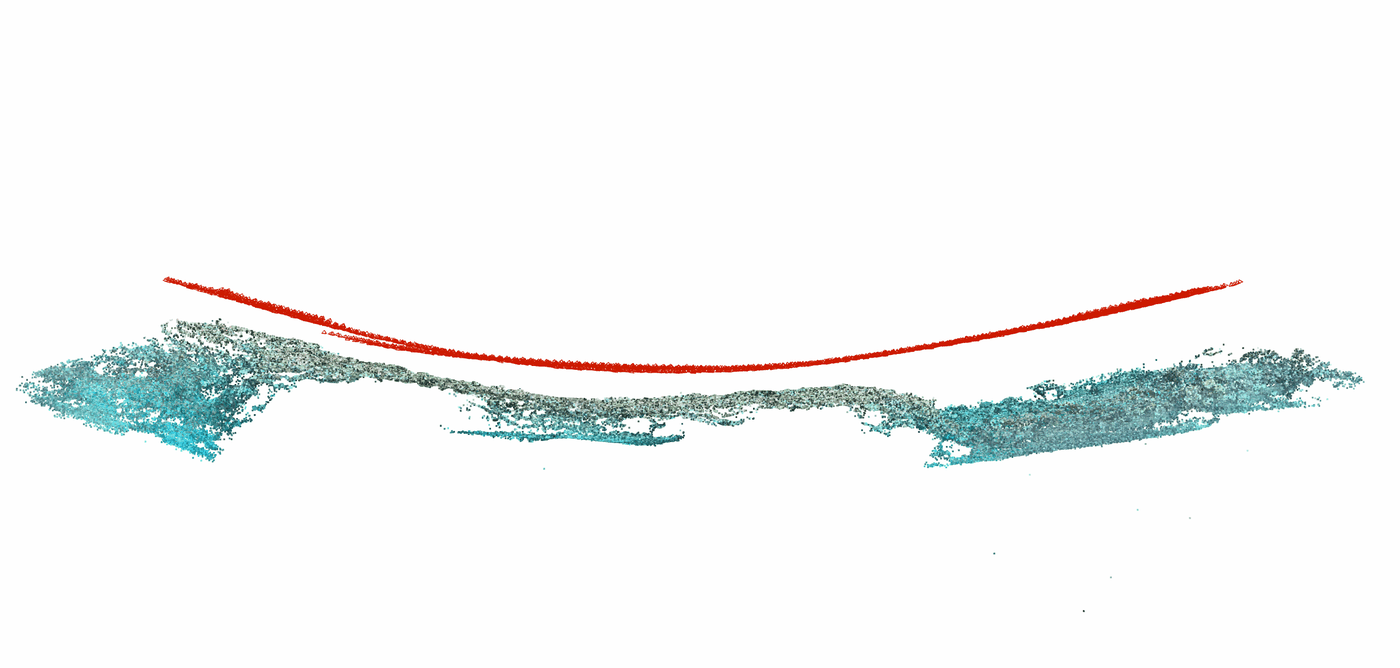} &
            \lizardpair{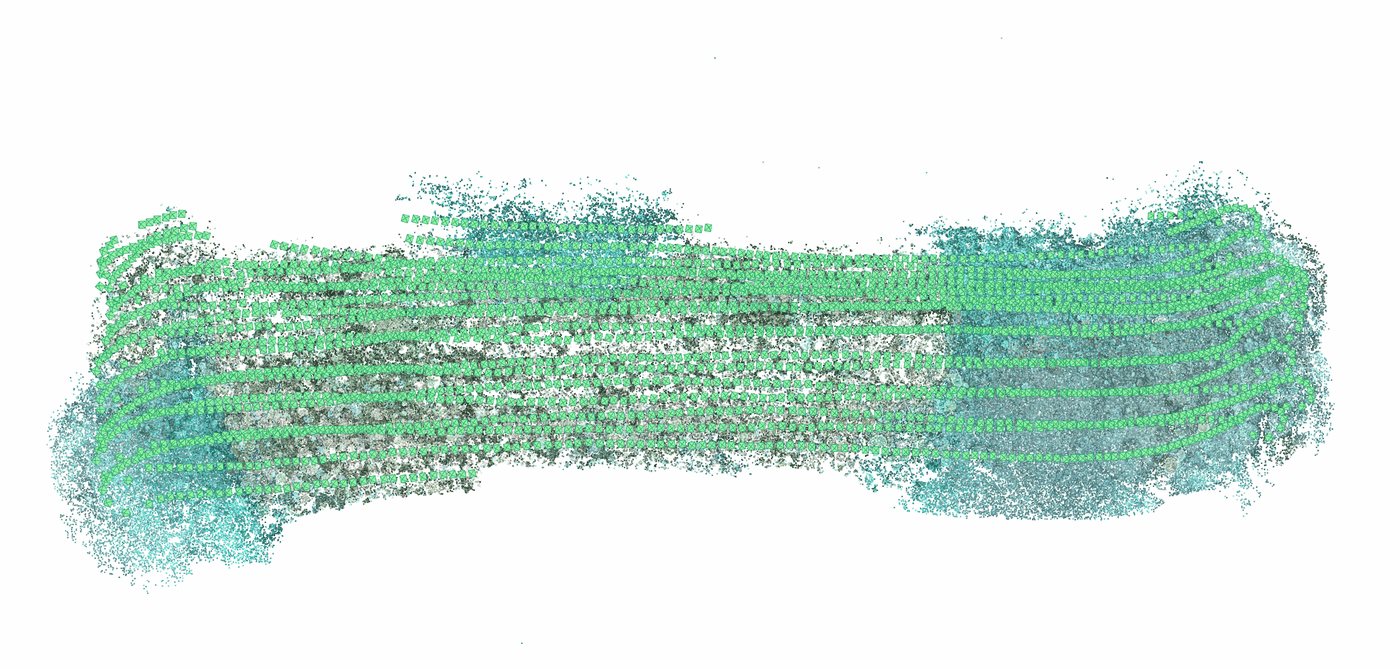}{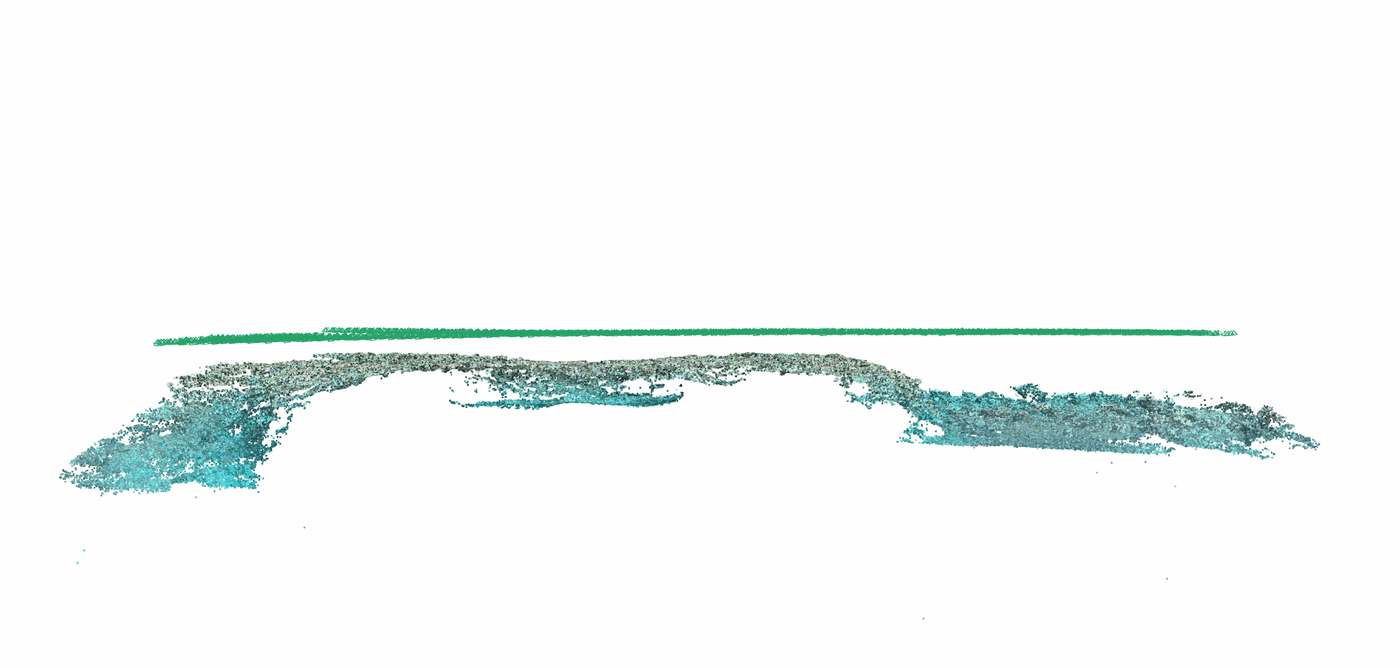} \\[0.3em]
            \hline\\[-0.8em]
            \lizardlabel{Mar'24 GP1} &
            \lizardpair{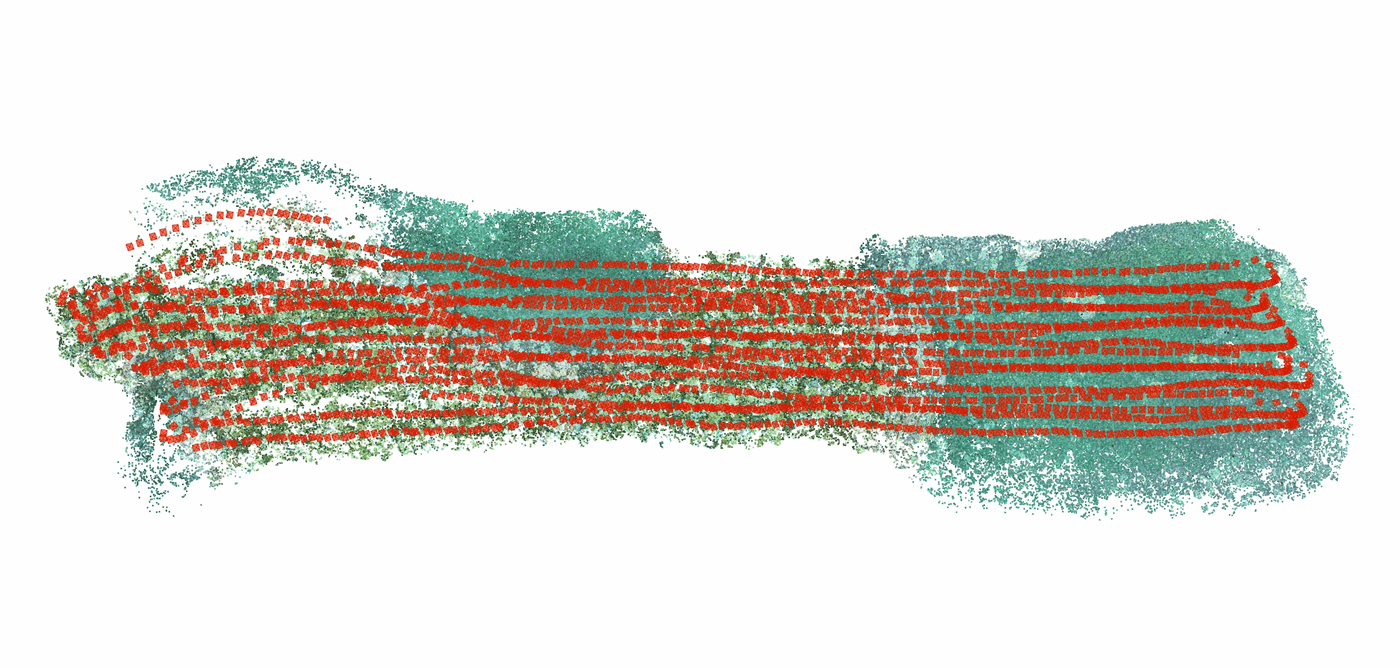}{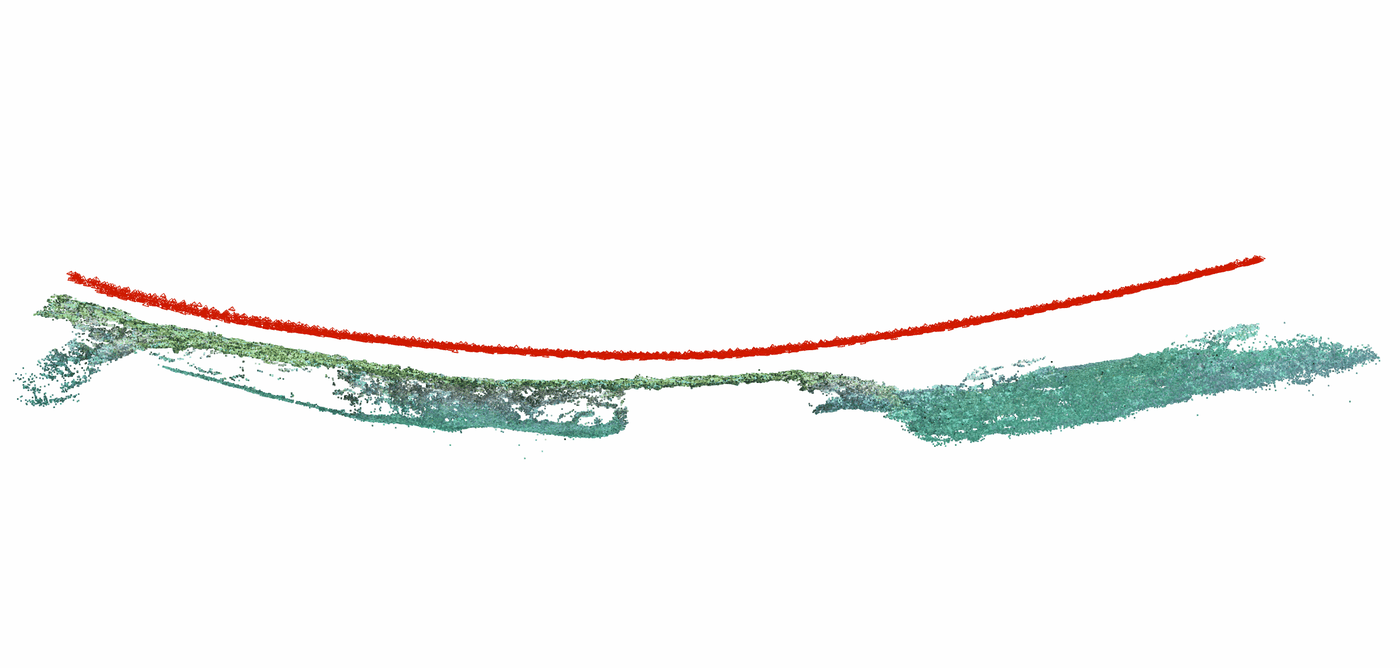} &
            \lizardpair{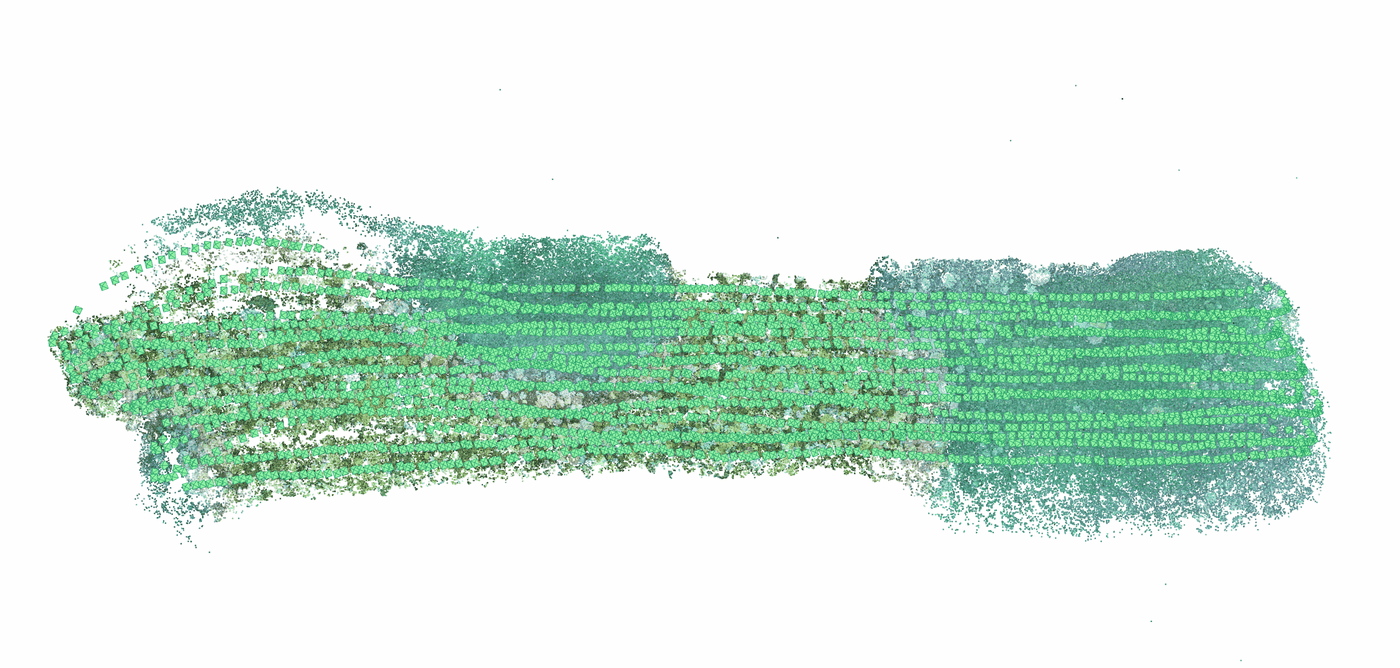}{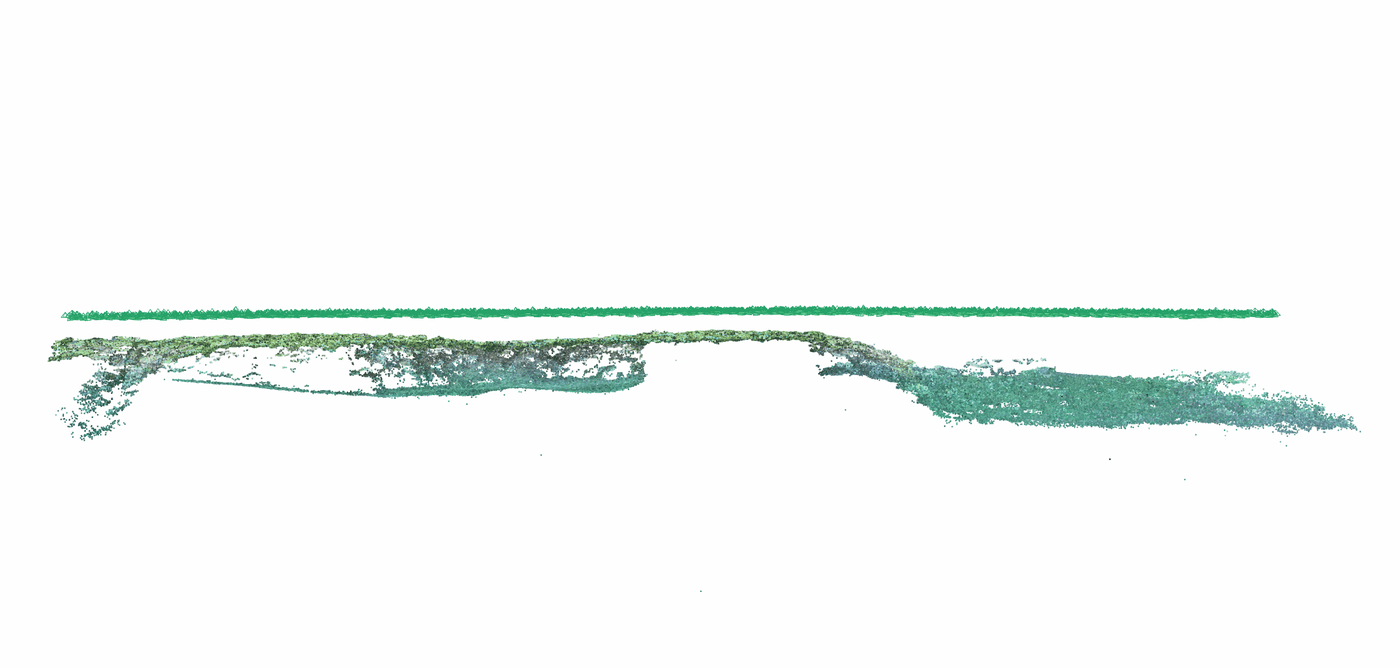} \\[0.3em]
            \hline\\[-0.8em]
            \lizardlabel{Mar'24 GP2} &
            \lizardpair{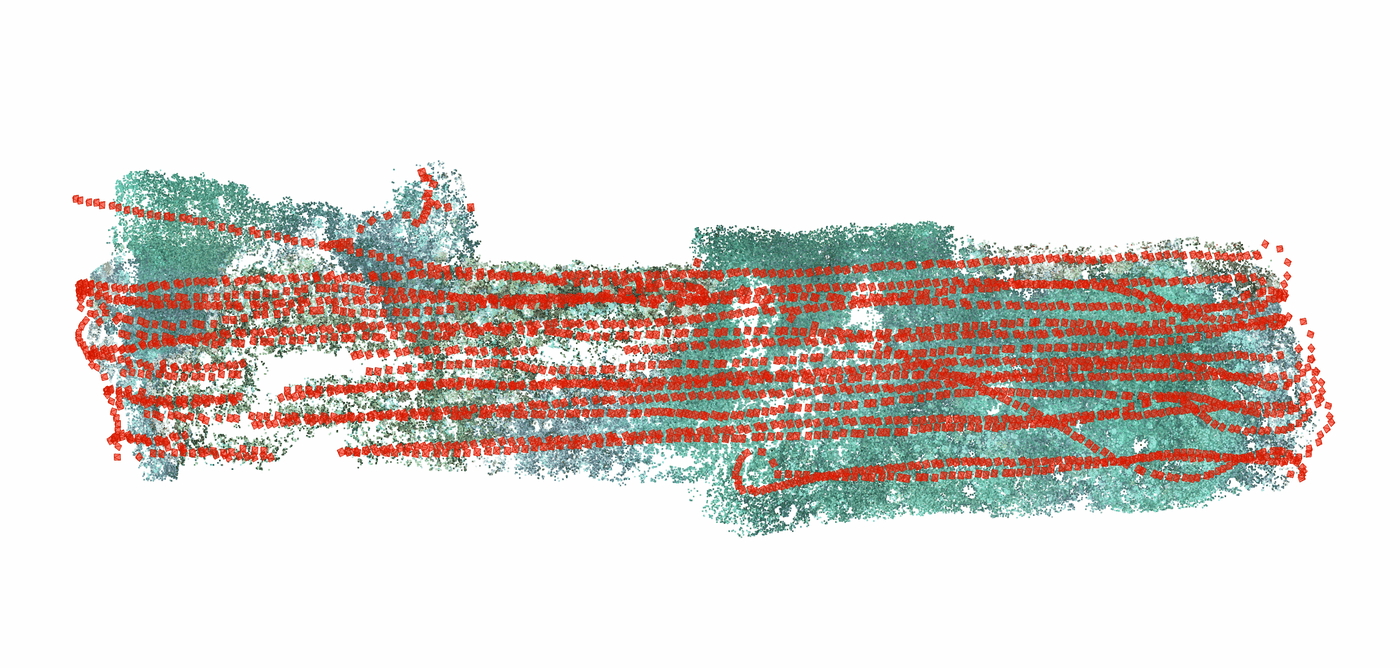}{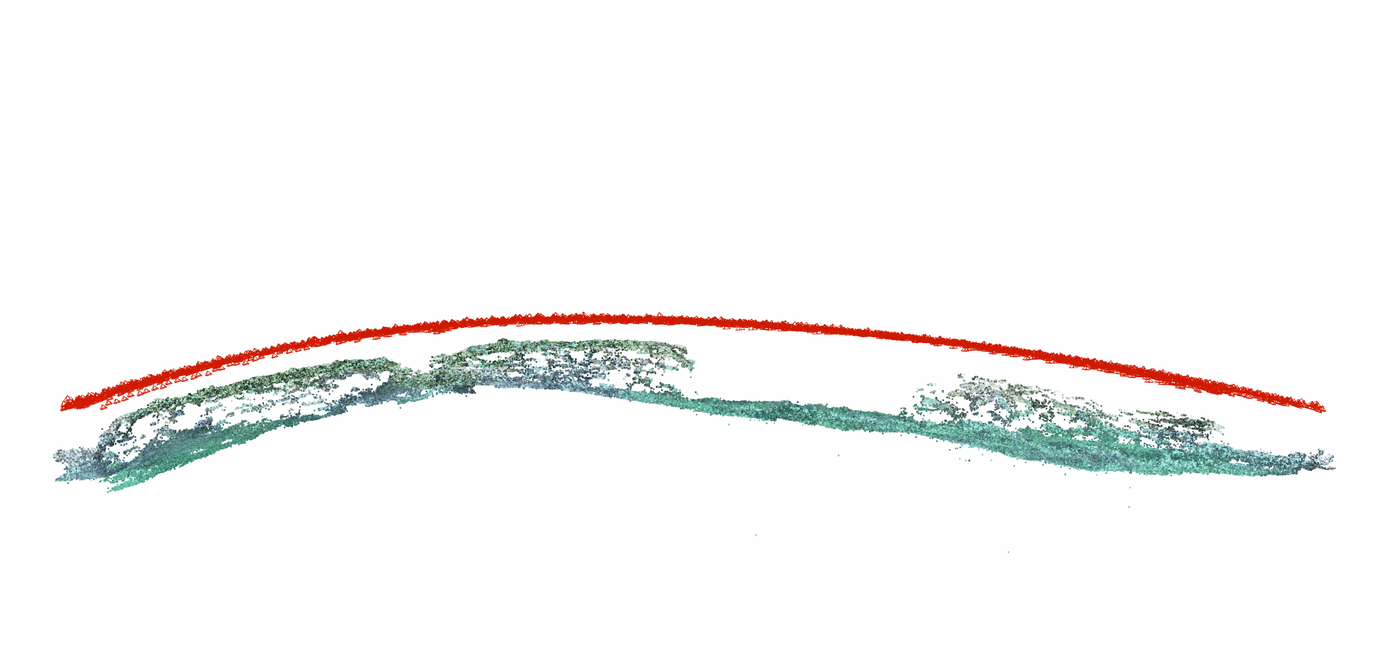} &
            \lizardpair{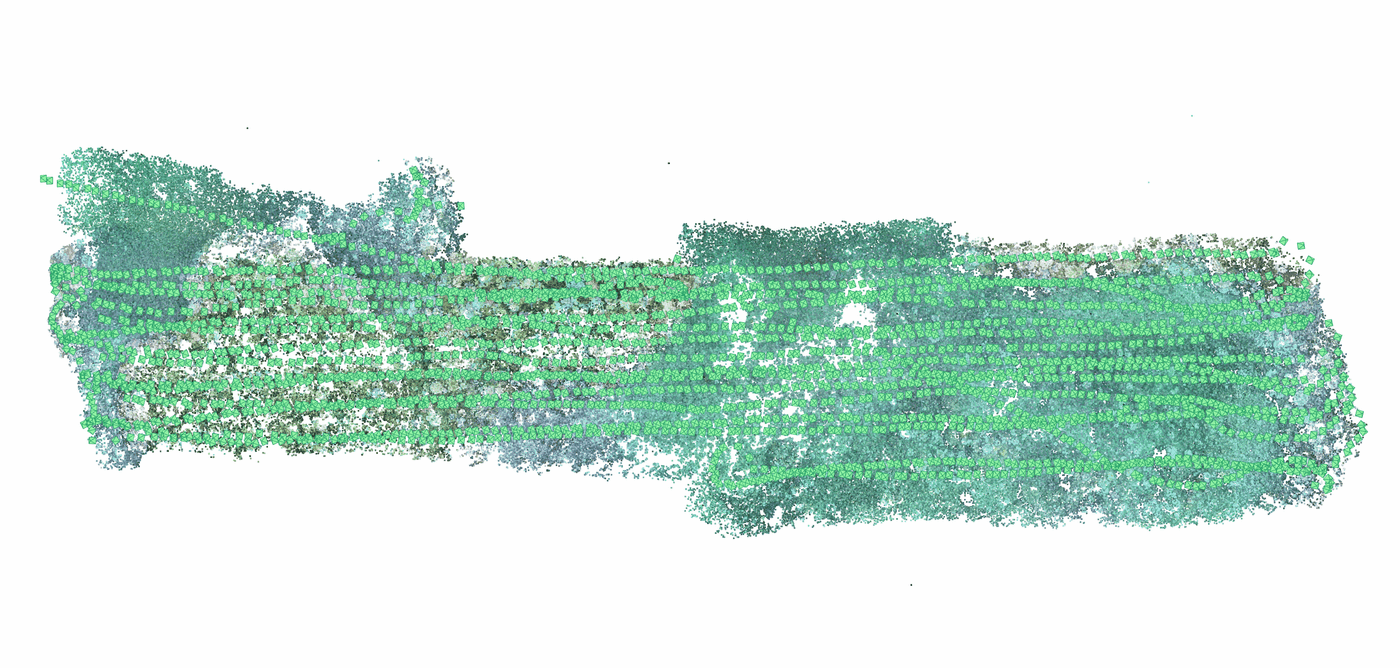}{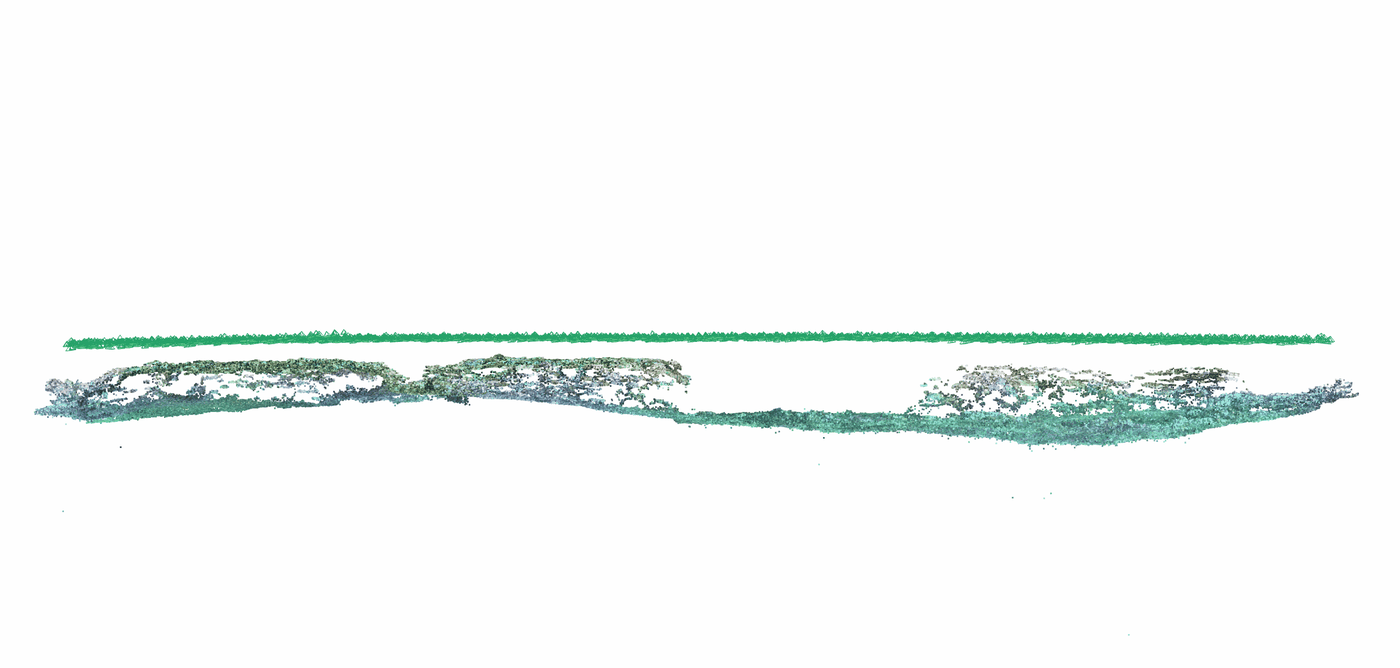} \\[0.3em]
            \hline\\[-0.8em]
            \lizardlabel{Sep'24 GP1} &
            \lizardpair{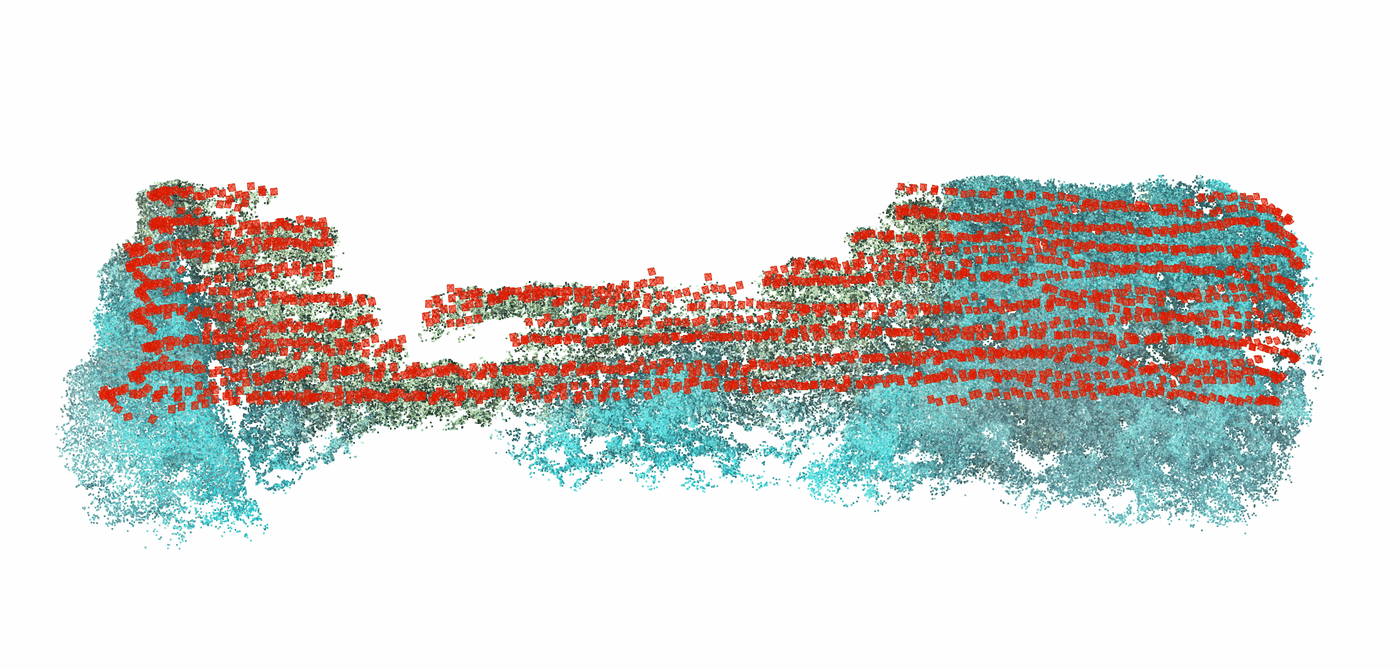}{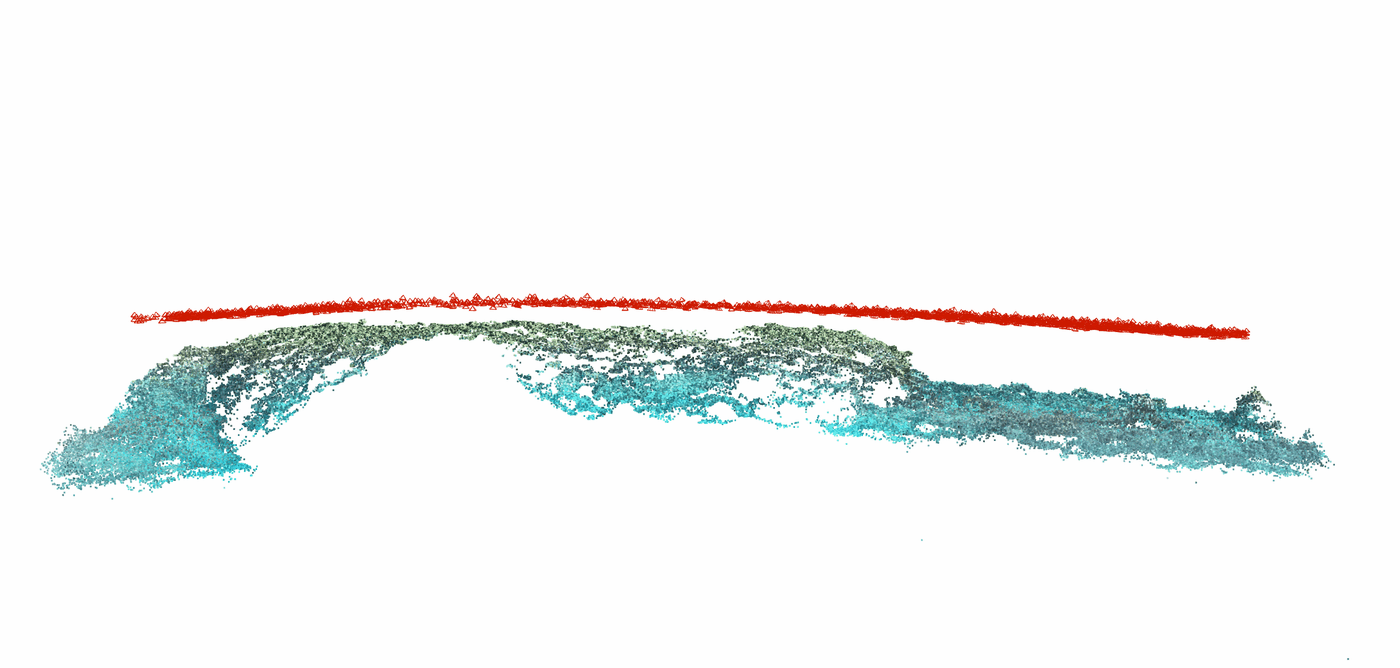} &
            \lizardpair{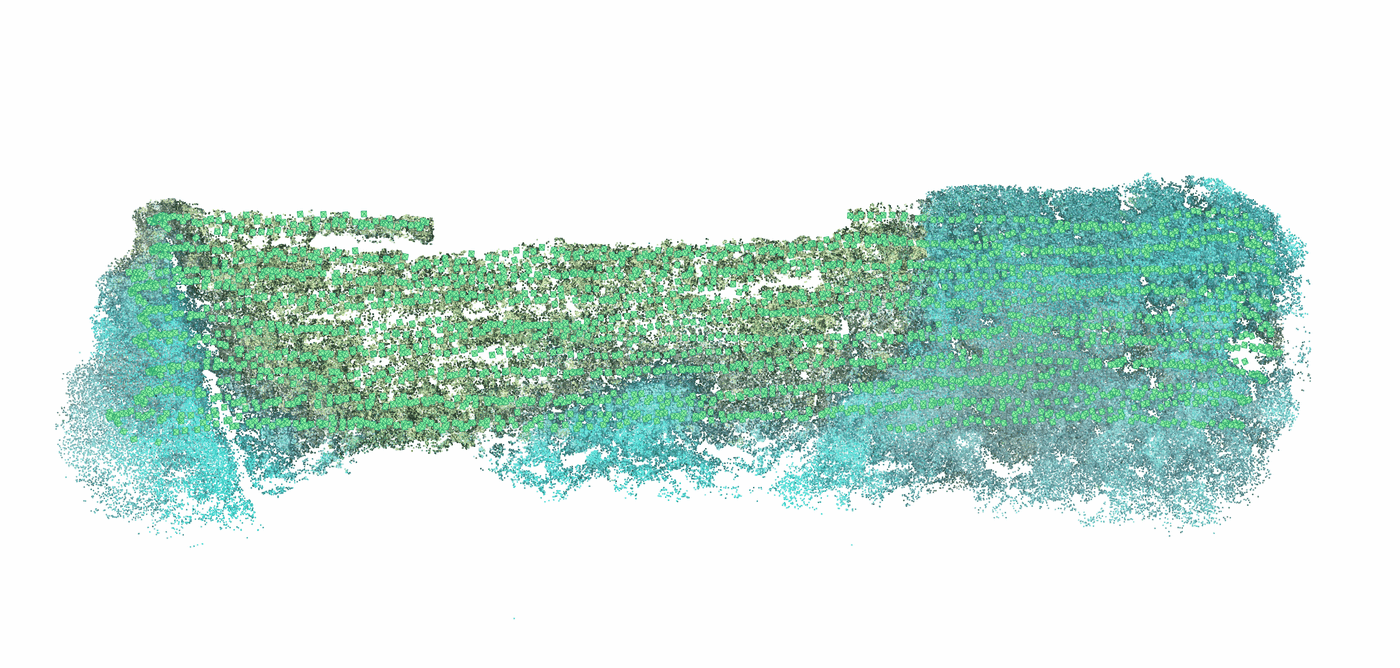}{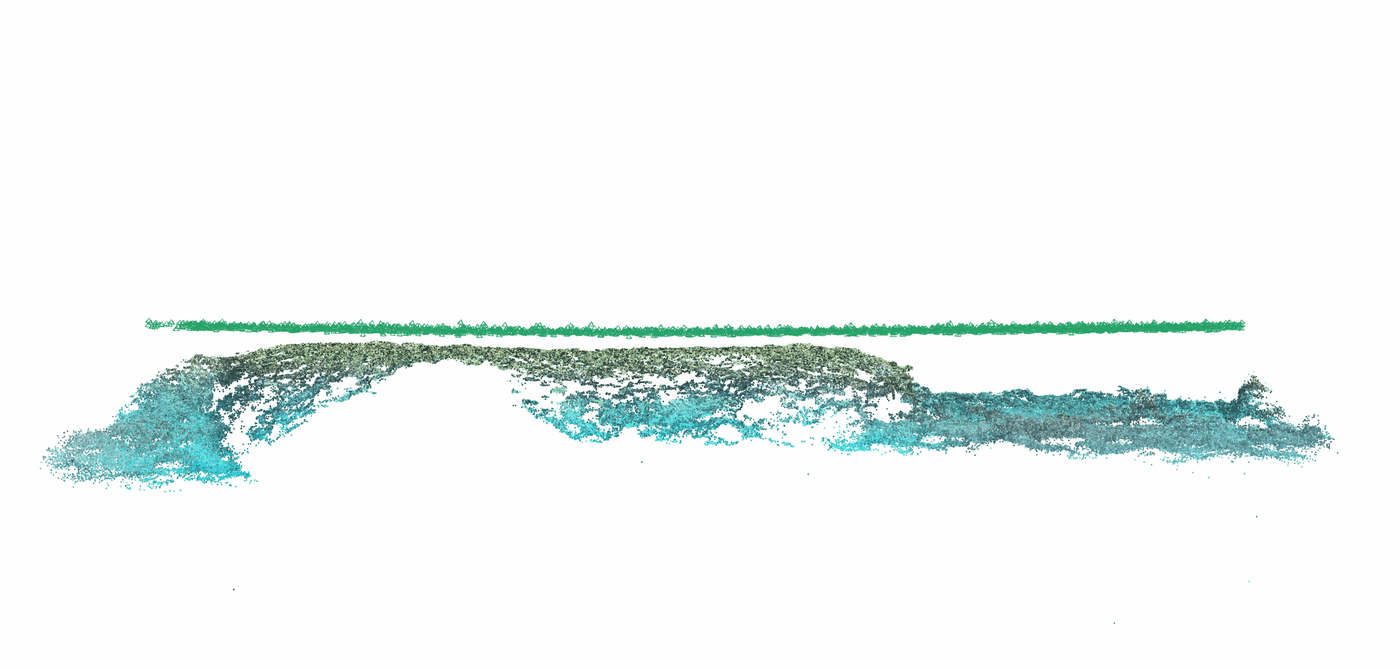} \\[0.3em]
            \hline\\[-0.8em]
            \lizardlabel{Sep'24 GP2} &
            \lizardpair{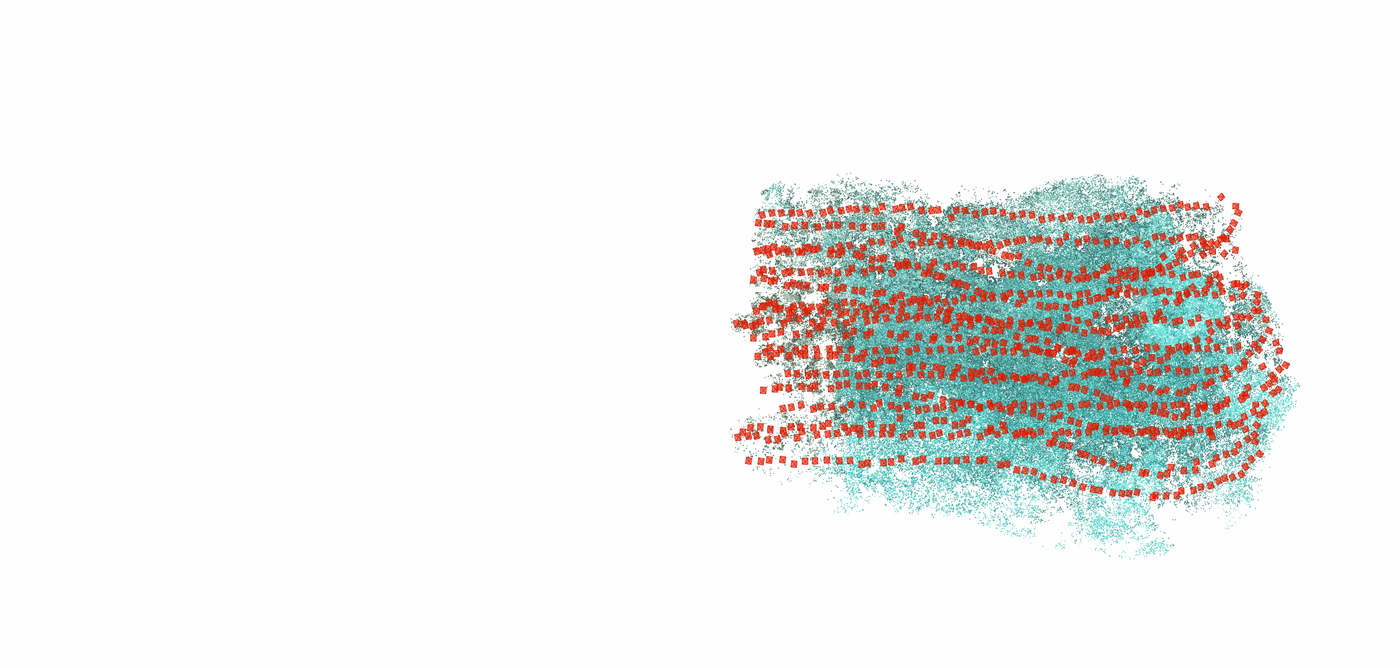}{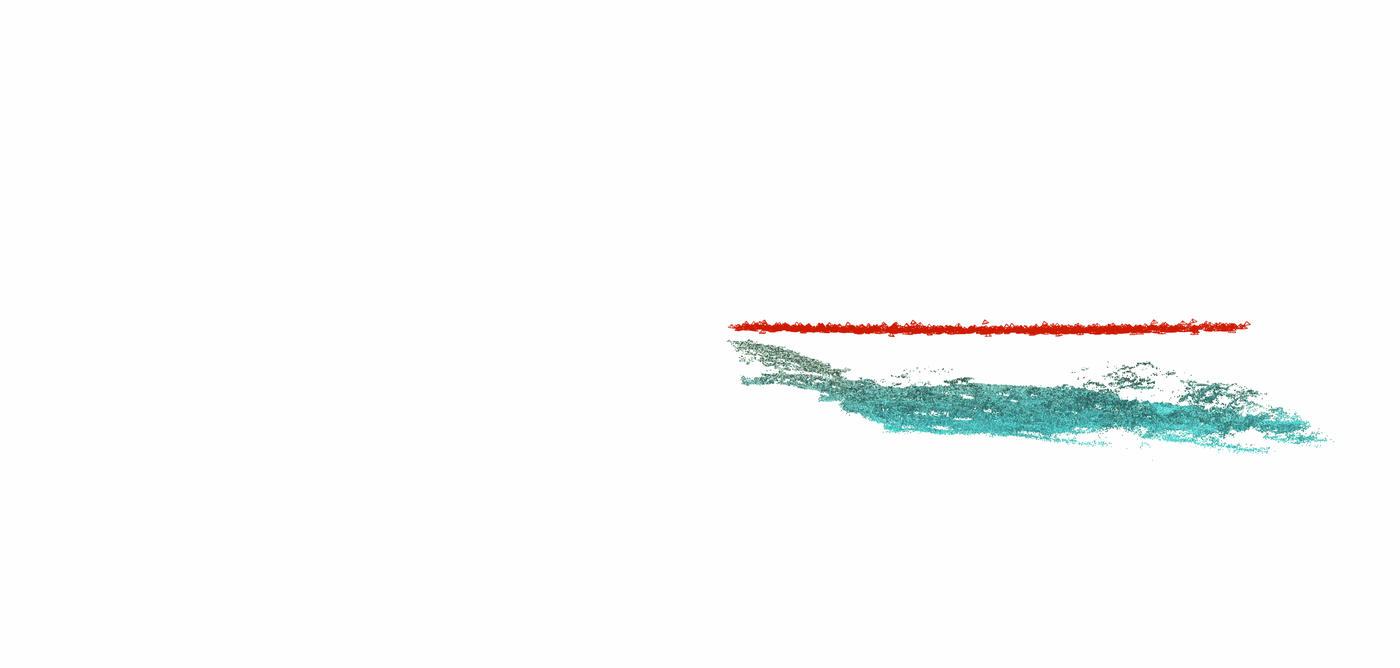} &
            \lizardpair{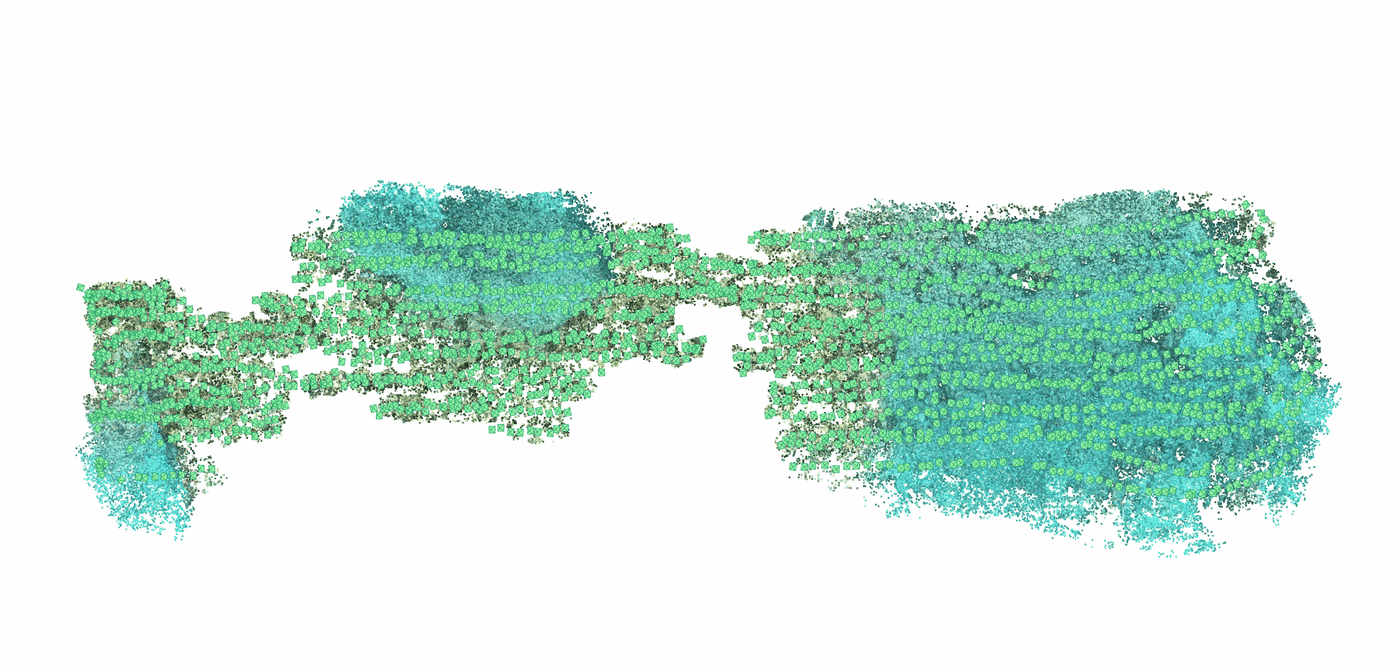}{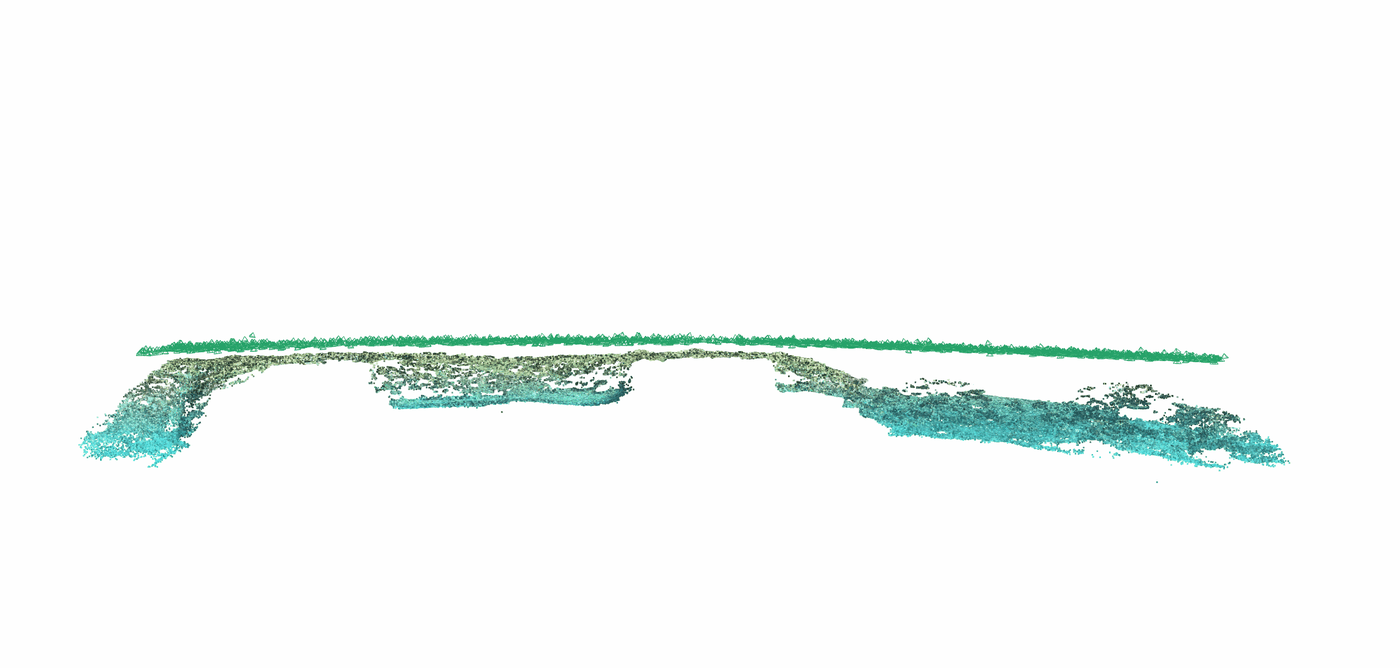} \\[0.3em]
            & \footnotesize COLMAP (ref)  &
            \footnotesize OURS + COLMAP \\
        \end{tabular}
        \label{fig:lizard-island}
    \end{subfigure}

    \vspace{0.8em}

    \begin{subfigure}[t]{\linewidth}
        \centering
        \caption*{\textbf{Malaysia}}
        \begin{tabular}{@{}c@{\hspace{0.01\linewidth}}c@{\hspace{0.02\linewidth}}c@{}}
            \malaysialabel{P3 S02} &
            \malaysiapair{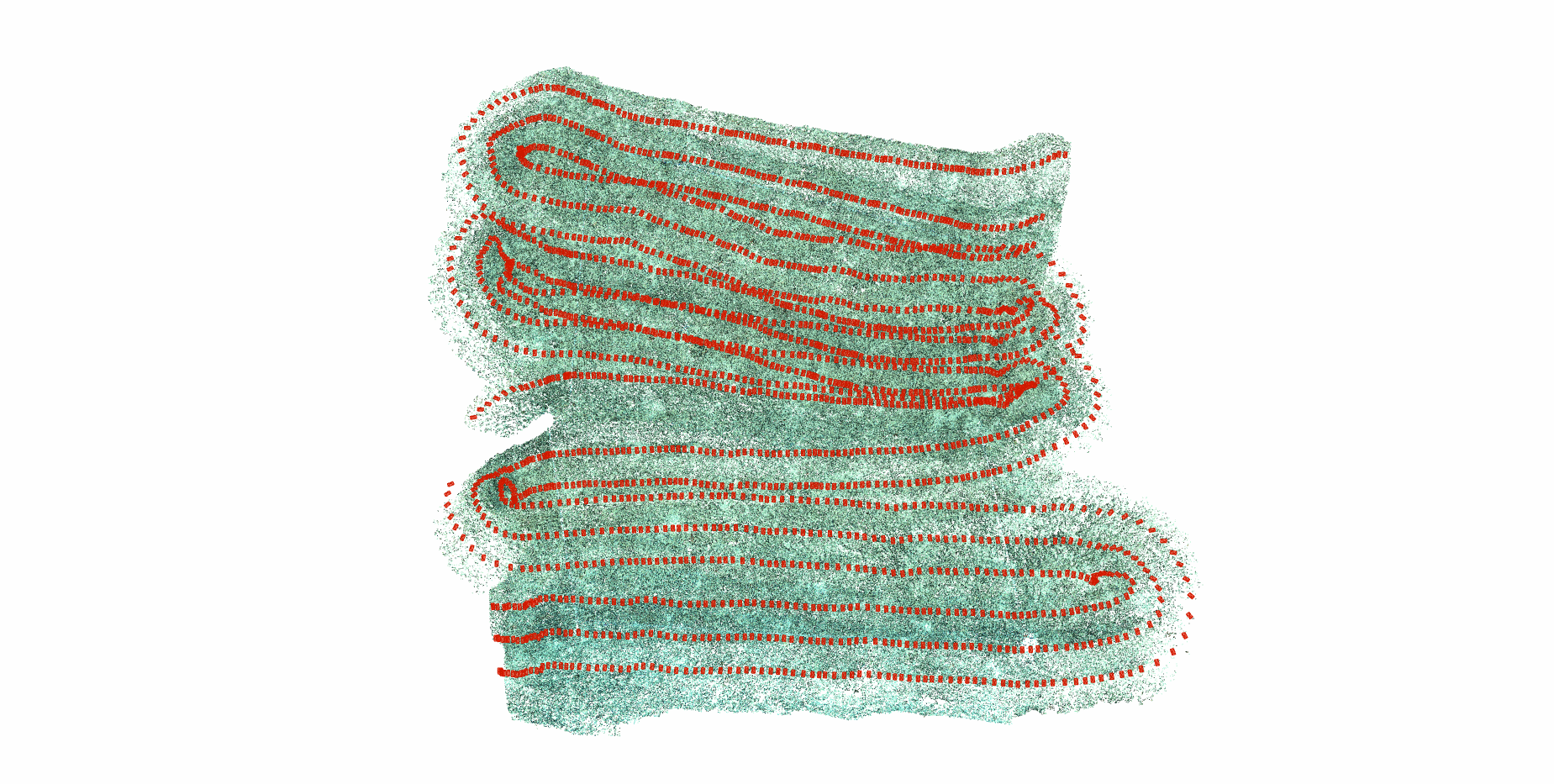}{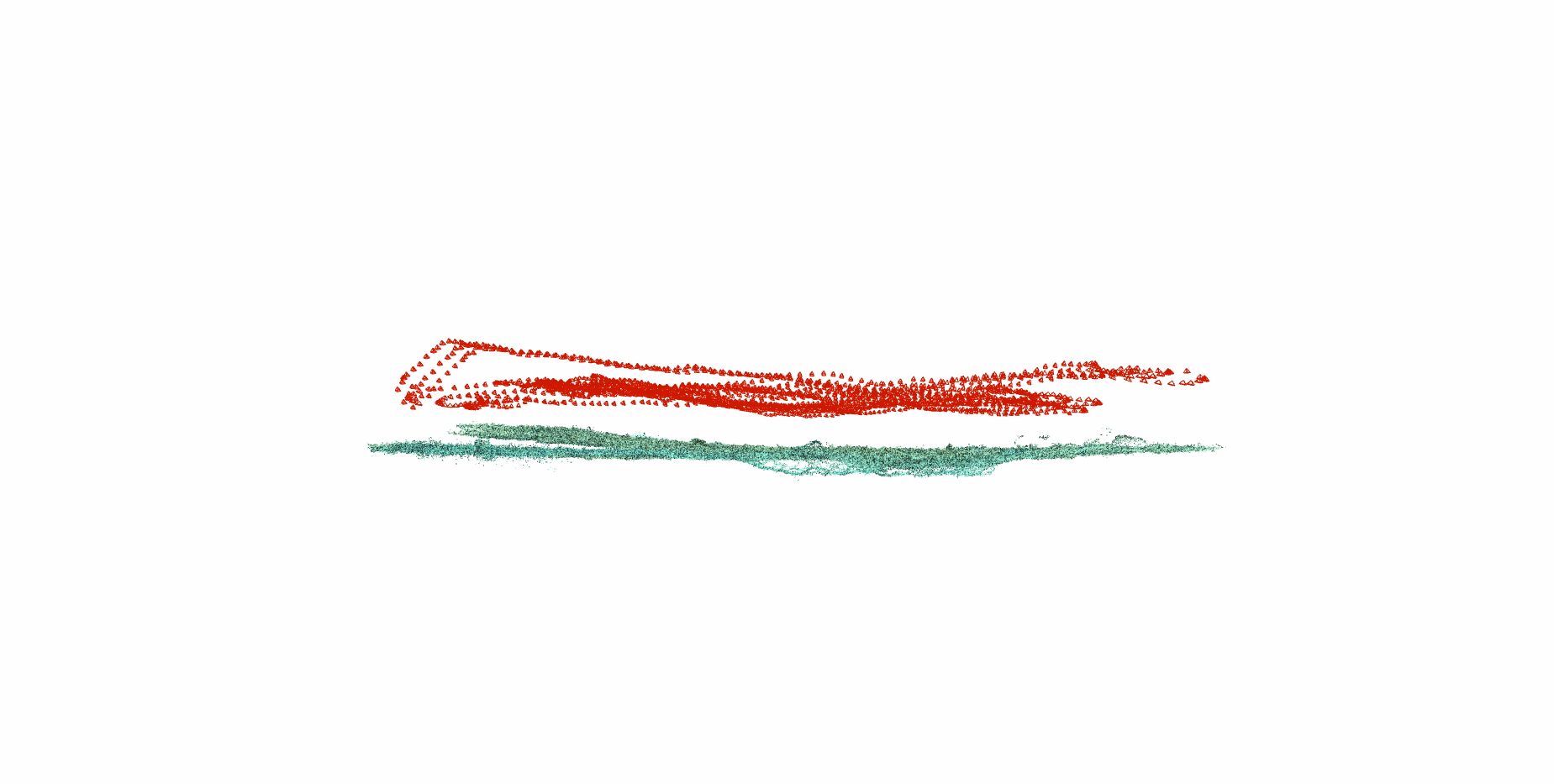} &
            \malaysiapair{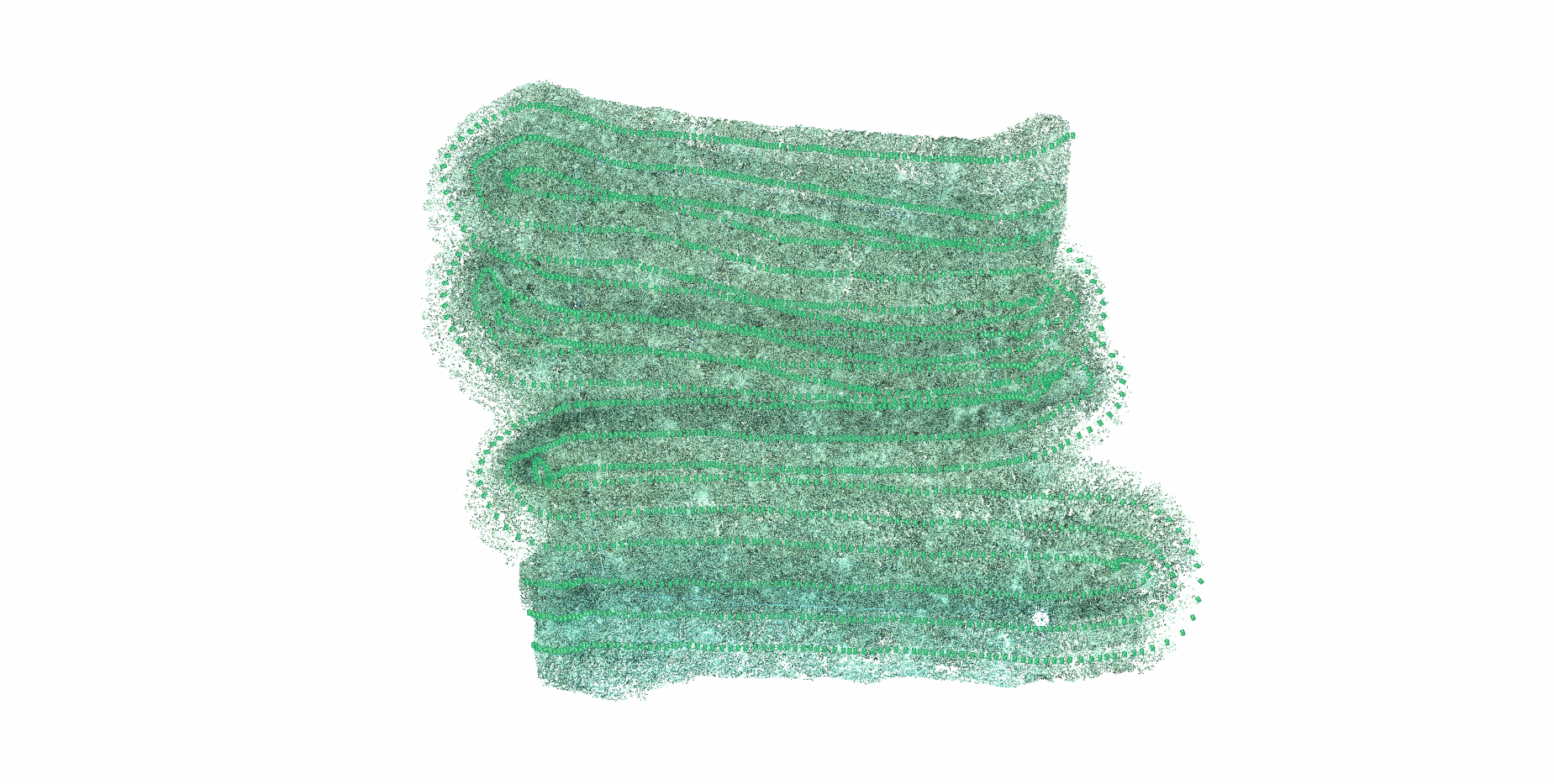}{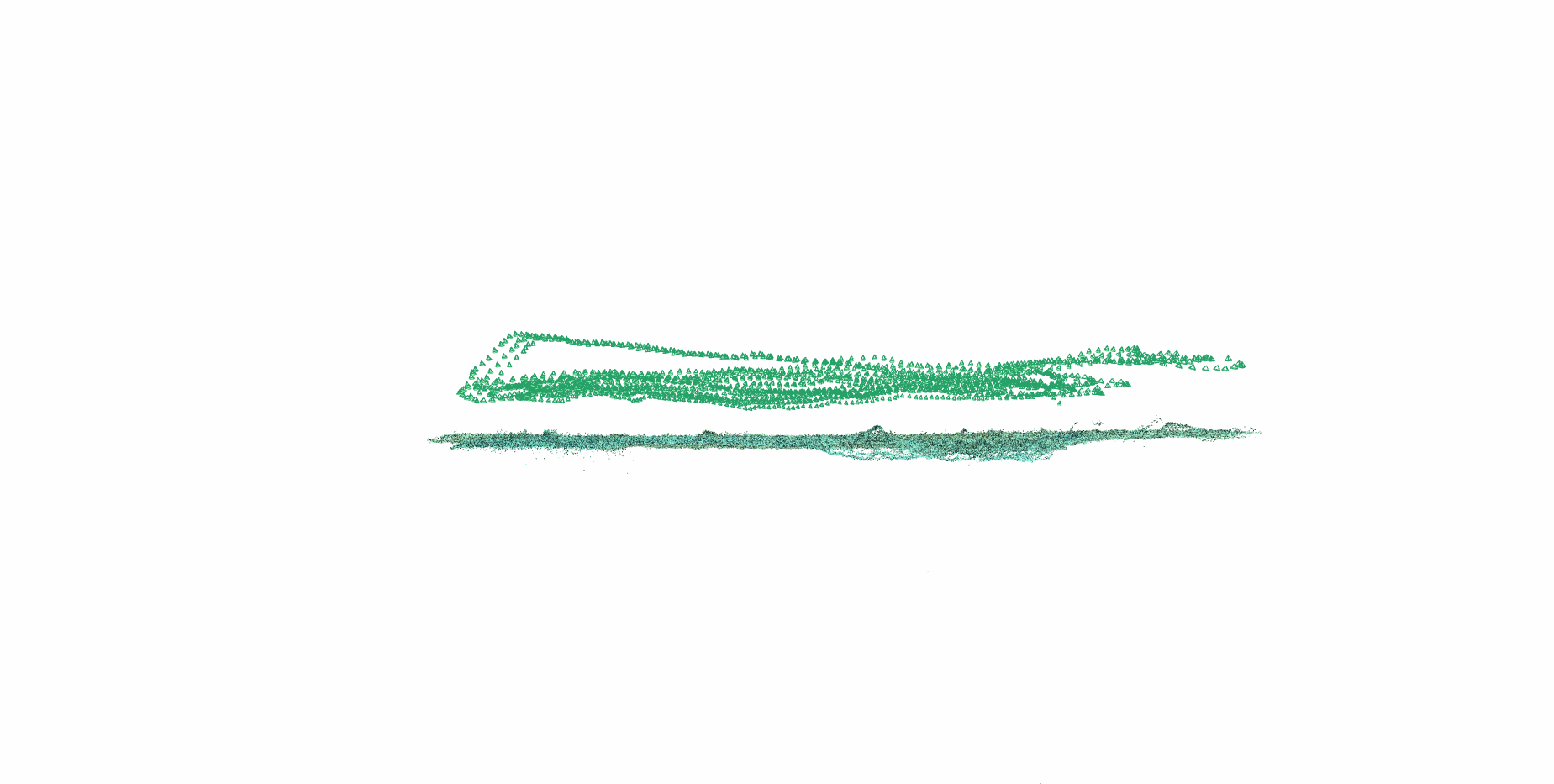} \\[0.3em]
            & \footnotesize COLMAP (ref)  &
            \footnotesize OURS + COLMAP \\
        \end{tabular}
        \label{fig:malaysia}
    \end{subfigure}

    \caption{\textbf{Comparison of 3D point clouds and camera trajectories} reconstructed using COLMAP (ref) and OURS + COLMAP across Lizard-Island and Malaysia datasets.  Note how our approach yields more complete trajectories (fewer camera gaps) and more geometrically consistent reconstructions (no bending).}
    \label{fig:lizard-malaysia}
\vspace*{-0.2cm}
\end{figure}

\textbf{Structure-from-Motion statistics.}
Even though absolute accuracy cannot be obtained, a systematic analysis of SfM statistics allows us to draw useful conclusions about the performance of the different baselines.

\emph{Registered image ratio (Reg.) and number of reconstructed 3D points (Pts):} ceteris paribus, a higher number of recovered camera poses and/or number of 3D points indicates a better fit of the model to the underlying 3D geometry.

\emph{Mean track length (Track):} the average number of images in which a reconstructed point is observed reflects how consistently features are matched across views.

\emph{Reprojection error (Reproj.):} smaller reprojection errors after bundle adjustment are indicative of a better fit of the model, but, as shown in Sec.~\ref{sec:results}, this does not necessarily translate to a more accurate reconstructed geometry.

\section{Results}
\label{sec:results}



We run all experiments using calibrated in-air intrinsics and standard flat-port parameters of $t_g = 2$\,mm glass thickness and $d_0 = 2$\,cm housing-to-camera distance. Results, reported in Table~\ref{tab:main_results} and Figure~\ref{fig:lizard-malaysia}, use the operational depth $Z_0$ obtained as detailed in the ablation study (Sec.~\ref{sec:ablation}).

\subsection{Geometric Fidelity via Structural Deformation}

Conventional COLMAP (fix) with fixed intrinsics always leads to catastrophic failure on both the Malaysia, and Lizard Island datasets. Refining the camera intrinsics and distortion parameters, COLMAP (ref), substantially reduces the curvature in Malaysia scenes, with $\kappa \in [-0.06, +0.19]$. These results are consistent with the fact that the scene structure in this dataset is generally planar, so refraction has an overall radial effect that can be modeled with radial distortion parameters. Consistently, on Lizard Island, where the scene is topographically varied and refraction produces a strong non-radial component, COLMAP (ref) remains comparatively larger, with $\kappa \in [-0.37, +0.31]$.

Refractive COLMAP, with $\kappa \in [+0.10, +0.69]$ in Malaysia and $\kappa \in [-0.45, +0.07]$ in Lizard Island, does not consistently outperform COLMAP (ref) with refined intrinsics, suggesting that explicitly modeling refraction hinders optimization convergence, likely causing it to settle in local minima.

In contrast, our image-space correction consistently produces the smallest structural curvature across the evaluated sequences, with $\kappa \in [-0.07, +0.08]$ in Malaysia and $\kappa \in [-0.02, +0.05]$ in Lizard Island sequences.

\subsection{Reconstruction Completeness and Model Fit}

\begin{figure}[t!]
    \centering
    \includegraphics[width=\columnwidth]{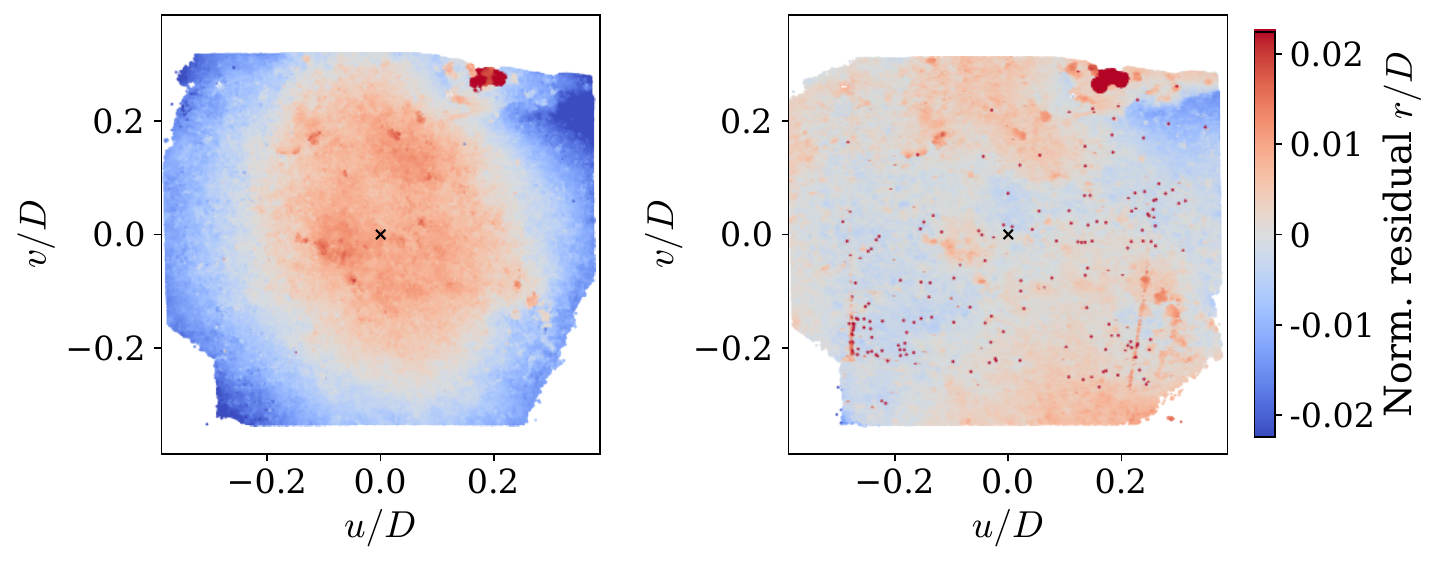}
    \caption{\textbf{Planarity deviation maps} for the Malaysia dataset p1\_s02 (left: Refractive COLMAP; right: Ours + COLMAP). Axes denote normalized in-plane coordinates; color shows normalized signed deviation from the fitted plane. The left panel shows a strong dome artifact, a hallmark of optimization convergence overfitting. Our method (right) eliminates this systematic pattern, leaving only deviations
    from the actual 3D structure.
    }
    \label{fig:residual_map}
\end{figure}
\begin{figure}[t!]
\centering
\begin{subfigure}[b]{0.49\linewidth}
\centering
\includegraphics[width=\linewidth]{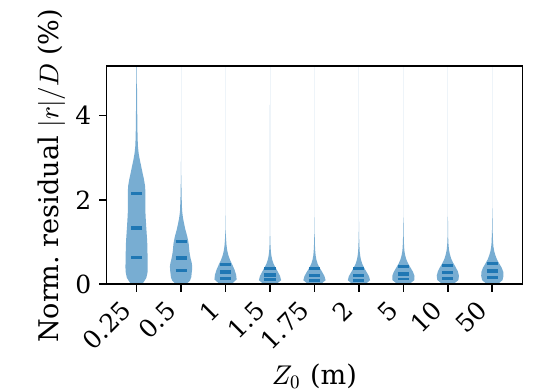}
\caption{Malaysia (p1\_s02)}
\label{fig:depth_ablation_malaysia}
\end{subfigure}
\hfill
\begin{subfigure}[b]{0.49\linewidth}
\centering
\includegraphics[width=\linewidth]{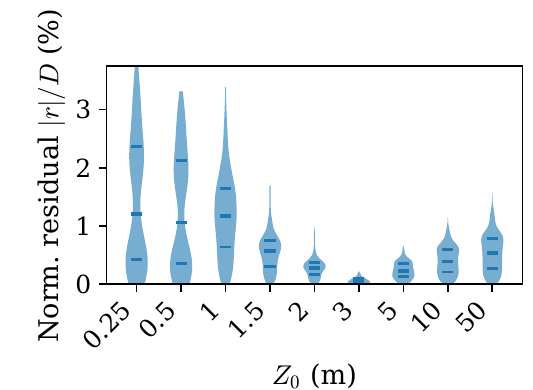}
\caption{Lizard Island (feb\_gp1)}
\label{fig:depth_ablation_lizard}
\end{subfigure}
\caption{\textbf{Depth ablation} to select the operational depth $Z_0$ used for refractive undistortion. We undistort images using a range of candidate values of $Z_0$ and reconstruct the scene for each. Each plot shows the resulting distribution of normalized planarity residuals across the sequence.}
\label{fig:depth_ablation}
\end{figure}

As shown in Table~\ref{tab:main_results}, our approach achieves the highest registered image ratio across all sequences with the highest geometric fidelity. As discussed in the previous section, COLMAP (ref) with refined distortion parameters consistently yields the lowest reprojection error on the Malaysia sequences, because the scene structure in this dataset is generally planar, so refraction has an overall radial effect that can be modeled with radial distortion parameters; however, this does not translate to the highest geometric fidelity.


Finally, another notable trend is that our approach produces the highest number of \textit{3D points} across 9/10 sequences and the highest \emph{mean track length} across all sequences. This indicates that our correction does not merely reconstruct more points at the cost of weaker constraints; rather, it recovers more 3D points while each is observed more consistently across views, better constraining the reconstruction.


\subsection{Generalization Across 3D Reconstruction and VSLAM}
A key benefit of our approach is that, by correcting refraction in image space, it allows seamless combination with any 3D reconstruction or VSLAM pipeline that assumes a perspective (pinhole) camera model, many of which are trained on in-air data. Table~\ref{tab:compat} shows how our approach converts catastrophic failure into standard operating performance for each system (see accompanying video).

\section{Ablation Studies}
\label{sec:ablation}


\newcommand{\hdrrot}{60}

\begin{table}[t]
\centering
\caption{\textbf{Generalization across 3D Reconstruction and VSLAM}. Trajectory planarity ($\kappa$) and number of registered images (Reg.) with our corrected inputs \emph{within} each backend.}
\label{tab:compat}
\resizebox{\linewidth}{!}{%
\begin{tabular}{lccccccc}
Input & \rotatebox{\hdrrot}{DA3~\cite{depthanything3}} & \rotatebox{\hdrrot}{MapAny.~\cite{keetha2026mapanything}} & \rotatebox{\hdrrot}{ORB3~\cite{campos2021orb}} & \rotatebox{\hdrrot}{AFV~\cite{fontan2024anyfeature}} & \rotatebox{\hdrrot}{DROID~\cite{teed2021droid}} & \rotatebox{\hdrrot}{DPVO~\cite{lipson2024deep}} & \rotatebox{\hdrrot}{MASt3R~\cite{murai2025mast3r}} \\
\midrule
Reg.$\uparrow$ & 1.00 & 1.00 & 0.20 & 1.00 & 1.00 & 1.00 & 1.00 \\
$\kappa \approx 0$ & -0.032 & -0.045 &  (-0.009)  & +0.005 & +0.029 & +0.006 & -0.176 \\
\bottomrule
\end{tabular}%
}
\end{table}


\begin{table}[t!]
\centering
\caption{\textbf{Runtime evaluation (ms / image).} Our method achieves the fastest runtime across all stages and reports the fastest total runtime, despite introducing the additional refractive undistortion step.}
\label{tab:runtime_results}

\resizebox{\linewidth}{!}{%
\begin{tabular}{lcccccc}
\toprule
Method & Undist.$\downarrow$ & Feat.\ Extraction$\downarrow$ & Feat.\ Matching$\downarrow$ & Mapping$\downarrow$ & Total$\downarrow$ \\
\midrule

COLMAP (refined) & -- & 7.69 & 1158.80 & 3499.74 & 4666.23 \\
Refractive COLMAP & -- & 8.42 & 1129.88 & 2934.84 & 4073.14 \\
Ours + COLMAP & 7.75 & 7.03 & 1015.05 & 2242.73 & \textbf{3272.56} \\
\bottomrule
\end{tabular}%
}
\end{table}



\textbf{Reference depth.}
As shown in Sec.~\ref{subsec:sensitivity}, $Z_0$ is strongly coupled with the flat-port parameter $d_0$; we therefore vary $Z_0$ directly and evaluate its effect on reconstructed geometry, using planarity as a proxy for geometric fidelity.

Fig.~\ref{fig:depth_ablation} shows that geometric fidelity is highest (curvature is minimized) at specific operational values $Z_0$. A limitation of this analysis is that we fit a single $Z_0$ per scene, whereas the true optimal depth varies per pixel with local scene depth. The value we identify is therefore an average that minimizes the aggregate curvature over the whole scene, rather than the locally optimal depth at every point (see Fig.~\ref{fig:residual_map}).




\textbf{Computational Performance.} Table~\ref{tab:runtime_results} shows that our approach speeds up feature matching (which includes geometric verification) and mapping, both by using a better-fitting model compared to COLMAP and a simpler model compared to Refractive COLMAP. We run all experiments on a desktop workstation with an  Intel Core i7-14700, 32GB RAM, and an NVIDIA RTX 4090.



\section{Conclusion and Future Work}


We presented an image-space refraction correction method for underwater 3D reconstruction with flat-port cameras. Our method corrects the refractive displacement by warping images to their equivalent perspective appearance prior to reconstruction. A key benefit is that this enables any standard 3D reconstruction or VSLAM pipeline to run unmodified.

Through this work we have identified current limitations. Specifically, we determine a single operational scene distance $Z_0$ by tuning it against prior knowledge of scene structure. However, this parameter should instead be obtained per pixel, either through stereo triangulation or recent monocular depth prediction models. More broadly, the field lacks an underwater dataset with highly accurate ground-truth geometry for 3D reconstruction.

{\small
\bibliographystyle{IEEEtran}
\bibliography{references, fontan}
}

\end{document}